\documentclass[final,5p,times,twocolumn]{elsarticle}

\usepackage{amssymb}
\usepackage{amsmath}
\usepackage{booktabs}
\usepackage{color}
\usepackage{array}
\usepackage{caption}
\usepackage{changepage}
\usepackage{multirow}
\usepackage{makecell}
\usepackage{threeparttable}
\usepackage{booktabs}
\usepackage{soul, color, xcolor}

\usepackage[colorlinks=true ]{hyperref}
\usepackage{orcidlink}
\usepackage{amsmath,amsfonts}
\usepackage{algorithm}
\usepackage{array}
\usepackage{pifont}
\usepackage{subfig}
\usepackage{multirow}
\usepackage{makecell}
\usepackage{graphicx}
\usepackage{subcaption}
\usepackage{listings}
\usepackage{colortbl}
\usepackage{hhline}
\usepackage{listings}
\usepackage{graphicx}
\usepackage{soul}
\usepackage{xcolor}
\sethlcolor{yellow} 
\usepackage{courier} 
\usepackage{float}
\usepackage{pifont}
\usepackage{algorithm}
\usepackage{algpseudocode}
\algrenewcommand\algorithmicrequire{\textbf{Input:}}
\algrenewcommand\algorithmicensure{\textbf{Output:}}

\definecolor{keywordcolor}{RGB}{200,0,200}
\definecolor{commentcolor}{RGB}{0,126,72}
\definecolor{stringcolor}{RGB}{163,21,21}
\definecolor{builtincolor}{RGB}{127,0,85}

\lstdefinestyle{pythonDM3D}{
	basicstyle=\fontfamily{pcr}\selectfont\footnotesize, 
	commentstyle=\color{commentcolor}\itshape,
	keywordstyle=\color{keywordcolor}\bfseries,
	stringstyle=\color{stringcolor},
	identifierstyle=\color{black},
	showstringspaces=false,
	language=Python,
	breaklines=true,
	frame=tb,
	rulecolor=\color{black},
	framerule=1.0pt,
	numberstyle=\tiny,
	xleftmargin=4pt,
	tabsize=4
}

\journal{Knowledge-Based Systems}

\begin{document}

\begin{frontmatter}



	\title{DM3D: Dynamic Mamba via Offset-Guided Feature Resampling \\ for Point Cloud Understanding}
	\author[label1]{Bin Liu
	}
	\ead{liubin@st.xatu.edu.cn}

	\author[label2]{Chunyang Wang
		\corref{cor1}}
	\cortext[cor1]{corresponding author}
	\ead{wangchunyang19@163.com}

	\author[label2]{Xuelian Liu
	}
	\ead{tearlxl@126.com}

	\author[label3]{Xuemei Li
	}
	\ead{lixuemei556677@163.com}

	\author[label1]{Ge Zhang
	}
	\ead{zhangge@st.xatu.edu.cn}


	\affiliation[label1]{
		organization={School of Opto-electronical Engineering, Xi'an Technological University},
		city={Xi'an},
		postcode={710021},
		country={China}}

	\affiliation[label2]
	{organization={Xi’an Key Laboratory of Active Photoelectric Imaging Detection Technology, Xi'an Technological University},
		city={Xi'an},
		postcode={710021},
		country={China}}

	\affiliation[label3]
	{organization={School of Mechanical and Control Engineering, Baicheng Normal University},
		city={Baicheng},
		postcode={137000},
		country={China}}

	\begin{abstract}
		State Space Models (SSMs) model long token sequences of point cloud with linear complexity, but require an unordered point cloud to be serialized. Existing methods mainly address this requirement by designing or learning a better token order. Even a well-constructed order, however, cannot preserve every local relation on an irregular 3D surface: a fixed sequence may still mix points that are close in index but distant in 3D or belong to different object parts.
		We propose DM3D, a dynamic Mamba architecture that preserves the base token order while adapting local feature support and state propagation.
		First,  according to local feature context, DM3D learns spatial and sequence offsets without constructing a global permutation. Then, spatial offsets adjust the sampling anchors in 3D space, whereas sequence offsets guide feature resampling within a local sequence window, which lets different slots draw from overlapping local supports while retaining their identities. 	This design preserves the global prior of the original traversal, allowing each token to aggregate a more suitable local context.
		Second, the state update is modulated by the 3D distance between points at adjacent sequence positions, thereby reducing information propagation when these points are spatially far apart.
		DM3D reaches 95.2\% accuracy on the ModelNet40, 93.3\% accuracy on the PB\_T50\_RS split of ScanObjectNN, and 84.8\% class mIoU on ShapeNetPart. Extensive experiments on benchmark datasets show that DM3D achieves strong and competitive performance, validating the effectiveness of local feature adaptation for point cloud understanding. The code is released on GitHub \url{https://github.com/L1277471578/DM3D}.

	\end{abstract}

	%


\end{frontmatter}




\section{Introduction}
\label{sec:intro}

3D point clouds provide a sparse representation of real-world geometry for autonomous driving \cite{TopNet, zhang_towards_2023, mambamos_2024}, robotics \cite{RobotCheng2022}, AR/VR \cite{ARLim2022, VR}, and immersive multimedia \cite{multimedia}. Their unordered and irregular structure, however, does not provide the native token order desired by sequential models \cite{pointnet, ptv1, dgcnn}. Mamba \cite{Mamba} is attractive to capture long-range dependencies among large numbers of point tokens,  because its computational complexity grows linearly with sequence length. However, this efficiency is obtained only after the point set has been serialized. The resulting order is therefore not a neutral preprocessing step: it determines which points are presented as local sequential context and which states are directly propagated into one another.

Existing point cloud Mamba methods primarily improve this interface by constructing a better sequence. Space-filling curves, consistent traversals, multi-path serialization, and spectral or voxel orderings increase spatial continuity or expose complementary neighborhoods \cite{PCM, PointMamba, GridMamba, HydraMamba, Pamba, SAST, ZigzagPointMamba}. Learned ordering methods instead use feature importance or semantic cues to rearrange point groups \cite{PoinTramba, DyReMamba}. These approaches change which tokens become neighbors in the sequence, but no 1D order can preserve every local relation on an irregular 3D surface. Near part boundaries, sparse regions, or folded surfaces, a local sequence window may still mix points that are distant in 3D or belong to different local structures. Measurements of sequence jumps and neighborhood preservation likewise show that serialization quality varies across traversal rules and locations \cite{Explor-lu2025}. Thus, even after choosing a strong base order, fixed local aggregation can still provide the current token with features from unsuitable points.

This limitation affects two successive stages of an SSM. First, unsuitable aggregation mixes information from spatially separated regions before the recurrent scan begins. Second, the recurrent update may carry that mixed representation to later tokens, particularly across a large geometric jump in the sequence. Geometry-aware point cloud SSMs improve local encoding or condition state transitions on geometric structure \cite{Mamba3D, strumamba3d, CloudMamba, PointSS}, but generally retain the feature sources supplied by each serialization window. Global or group-level reordering changes token adjacency, yet it does not provide a local, token-specific adjustment after the base traversal has been selected. These observations motivate a complementary strategy: retain the base order, adapt the features aggregated by each token within a bounded window, and reduce state propagation where large spatial discontinuities remain.

\begin{figure}[t]
	\centering
	\includegraphics[width=0.48\textwidth]{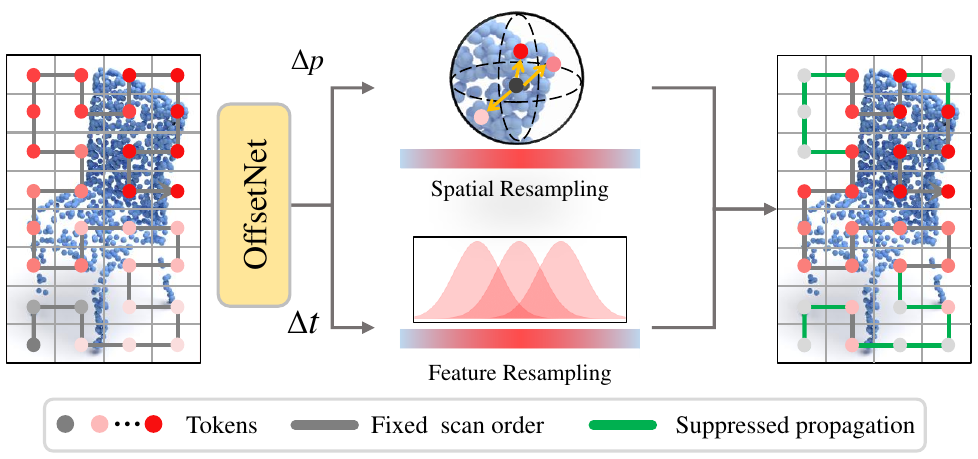}
	\caption{On the left, the context created by the fixed token order is not always consistent with the spatial and feature context of the current point. On the right, greater consistency between spatial neighborhoods and the aggregated feature context. Green lines denote that the feature propagation across discontinuities is suppressed.
	}
	\label{fig1}
\end{figure}

We therefore propose \textbf{DM3D (Dynamic Mamba for 3D point clouds)}, which preserves the base token order but changes the local features read by each token (see Fig.~\ref{fig1}). It first predicts spatial and sequence offsets from local feature differences. These offsets move a spatial sampling anchor and shift a bounded sequence-domain center, so each fixed slot receives a locally resampled feature rather than being moved to a new position. We refer to the sequence-domain operation as Offset-Guided Feature Resampling (OFR). Local resampling cannot remove every discontinuity, so it uses the distance between consecutive deformed anchors to scale the Mamba step size, thereby increasing the attenuation of the historical state across larger gaps. The dynamic path is integrated with two standard SSM paths to retain sequence context beyond the local candidate window. In this way, DM3D complements the global spatial prior of an existing traversal, avoids the constraints of global assignment, and preserves linear complexity for fixed candidate sizes.

Across object classification, few-shot learning, part segmentation, and scene semantic segmentation, DM3D remains competitive with recent point cloud models. The results show that base ordering and local feature adaptation are complementary, with the former providing a global spatial prior and the latter adjusting the context where that prior is locally insufficient.
Our contributions are summarized as follows:

\begin{itemize}
	\item We propose DM3D, unlike global reordering or permutation-based methods, that preserves the base token order while adapting both the local features read at each sequence slot and the state propagated between adjacent slots.

	\item We introduce  a dynamic path that first resamples local features at fixed sequence slots and then makes state propagation responsive to the geometric gaps that remain. This design addresses both the information entering the SSM and its subsequent propagation without learning a new global order, and retains linear complexity.

	\item Extensive experiments on benchmark datasets demonstrate that DM3D achieves highly competitive performance across classification, few-shot learning, and part segmentation tasks. More experiments further verify the respective roles of local feature resampling and geometry-aware state propagation.

\end{itemize}

\section{Related Work}
\label{sec:Related Work}

\subsection{Deep Learning for Point Clouds}
Deep point cloud models must jointly capture local geometry and broader contextual structure in an unordered set. PointNet \cite{pointnet} established direct set processing through shared point-wise mappings and symmetric aggregation, while PointNet++ \cite{PointNet2} introduced hierarchical local neighborhoods. Subsequent methods improve this local-to-global modeling through stronger point encoders and training recipes, including PointNeXt \cite{PointNeXt}, or through attention-based interaction, such as PCT \cite{PCT} and Point Transformer V3 \cite{ptv3}. Point-BERT \cite{Point-BERT} further shows that masked representation learning can provide transferable point cloud features. These methods provide effective feature extractors, but do not directly resolve how a serialized state space model should correct local feature context when its base traversal is only an approximate spatial prior.

\subsection{Serialization and Mamba for Point Cloud}
Point cloud serialization converts an unordered set into a sequence on which an SSM can operate. PointMamba \cite{PointMamba} uses space-filling curves, while PCM \cite{PCM} combines consistent traversals and order prompts to expose complementary spatial neighborhoods. Pamba \cite{Pamba}, VoxelMamba \cite{voxelmamba}, and GridMamba \cite{GridMamba} extend this idea through multi-path, voxel-based, or multiple space filling curve. SAST \cite{SAST} constructs a spectral traversal, and ZigzagPointMamba \cite{ZigzagPointMamba} uses a structured zigzag scan to improve spatial continuity during representation learning. These methods differ in how the base sequence is built, but each supplies a discrete traversal before the sequential computation.

A second line makes ordering task-adaptive. PoinTramba \cite{PoinTramba} ranks point groups by learned importance and forms bidirectional importance-aware sequences, whereas DyReMamba \cite{DyReMamba} combines semantics-driven dynamic reordering with bidirectional state modeling. Both methods alter the global or group-level order presented to the SSM. Their common objective is to construct a more suitable sequence before or during state-space modeling; whether the local feature support can instead adapt while the base order remains fixed is a separate question.

Beyond serialization and reordering, several point cloud SSMs improve the sequential model itself by making it more responsive to geometry. Mamba3D \cite{Mamba3D} enhances local features before sequence processing, StruMamba3D \cite{strumamba3d} introduces spatial states and state-wise updates, and CloudMamba \cite{CloudMamba} uses grouped selective state spaces to improve geometric perception. LFE-PointMamba \cite{LFE-PointMamba} enhances local geometry with multi-scale features and non-causal grouped convolution, while dynamic Hilbert-curve rearrangement captures global context. PointSS \cite{PointSS} introduces a Global Geometry-Aware Mechanism to exchange explicit geometric priors within serialization windows, and its Adaptive Scale-Decoupled SSM generates geometry- and scale-dependent state transitions.

In contrast to these architectural or ordering adaptations, DM3D retains the base traversal and learns token-specific feature support in both spatial and sequence domains before regulating state propagation across geometric discontinuities.

\subsection{Adaptive Feature Resampling}
Learned resampling adapts the feature support rather than the sequence itself. Deformable convolution and attention move sampling locations on regular grids \cite{dcn, dat}, whereas KPConv and related point operators adapt spatial kernels to irregular geometry \cite{KPConv, ptv2, 3d-gcn, PointConv}; DefMamba \cite{defmamba} additionally changes the scanning path for image SSMs. Differentiable sorting methods, including NeuralSort \cite{NeuralSort}, SoftSort \cite{SoftSort}, and Sinkhorn \cite{Sinkhorn}, instead relax rankings or permutations through structured assignment matrices.

DM3D makes a complementary choice, using offset-guided spatial and sequence-domain resampling to adapt the local features read at each slot. This retains the broad spatial structure of an existing traversal, avoids a second global ordering stage, and remains linear in sequence length for a fixed candidate size.

\section{Method}
\subsection{Preliminaries}

\begin{figure*}[t]
	\centering
	\includegraphics[width=0.94\linewidth]{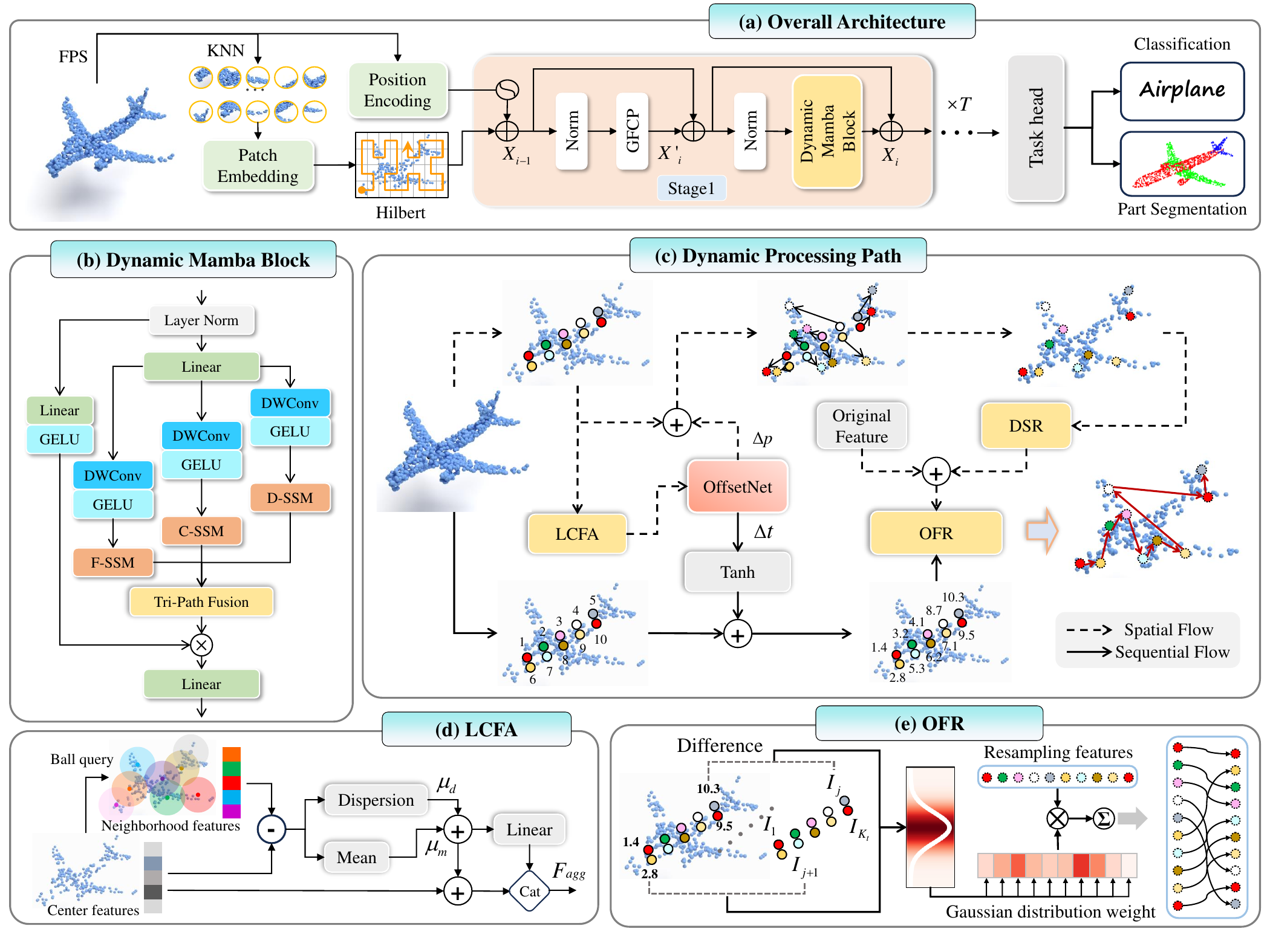}
	\caption{{Overview of DM3D.} {(a) Overall architecture} with the embedding, encoder, and task head. {(b) The Dynamic Mamba Block (DMB)} contains the proposed dynamic path, the forward SSM from PointMamba \cite{PointMamba} and the channel-flipped SSM from Mamba3D \cite{Mamba3D}. {(c) Dynamic processing path} predicts spatial and sequential offsets via OffsetNet, enabling offset-guided  resampling. ``Spatial flow'' and ``Sequential flow'' indicate the two resampling domains. {(d) LCFA} provides local contextual cues for the offset network. {(e) OFR} resamples features while keeping the output at its original sequence slot.
		{Symbols:} $Cat$ denotes concatenation along the channel dimension, $\odot$ element-wise multiplication, $\circleddash$ subtraction, $\oplus$ residual addition, $\otimes$ matrix multiplication, and $\sum$ summation.
	}
	\label{overview}
\end{figure*}

State Space Models (SSMs) represent a sequence through recurrent transitions of a latent state. Structured SSMs (S4) \cite{Gu2022} use Zero-Order Hold (ZOH) discretization with a step size $\Delta$, yielding
\begin{equation}
	\bar{A}=\exp(\Delta A), \quad
	\bar{B}=(\Delta A)^{-1}\left(\exp(\Delta A)-I\right)\Delta B
\end{equation}
\begin{equation}
	{{h}_{t}}=\bar{A}{{h}_{t-1}}+\bar{B}{{u}_{t}}, \quad {{y}_{t}}=C{{h}_{t}}
\end{equation}
where $\bar{A}$ and $\bar{B}$ are the discretized counterparts of the continuous state transition matrix $A$ and input projection matrix $B$, respectively. $C$ denotes the output projection matrix, while $u_t$, $h_t$, and $y_t$ represent the input, hidden state, and output at time step $t$, respectively.

Mamba \cite{Mamba} further makes $\Delta$, $B$, and $C$ input-dependent through its selective scan, allowing the state transition and input contribution to vary across tokens.

\subsection{Overview}
DM3D follows the ViT-style point cloud backbone used by PointMamba \cite{PointMamba} and Mamba3D \cite{Mamba3D}; Fig.~\ref{overview}(a) shows the complete pipeline.

Given an input point cloud, Farthest Point Sampling (FPS) and K-Nearest Neighbor (KNN) grouping produce $N$ local groups of $K$ points. A lightweight PointNet embeds these groups as $F\in\mathbb{R}^{N\times D}$, with group centers $P\in\mathbb{R}^{N\times 3}$. An MLP maps the center coordinates to position embeddings $P_{emb}\in\mathbb{R}^{N\times D}$, which are added to the group features. We then prepend a classification token $F_{CLS}\in\mathbb{R}^{1\times D}$ and serialize the point tokens with a Hilbert curve, forming the initial sequence $X_0\in\mathbb{R}^{(N+1)\times D}$.
Each encoder stage contains Geometry-Feature Coupled Pooling (GFCP) from HyMamba \cite{Hymamba} for local enhancement, followed by the proposed Dynamic Mamba Block (DMB). The $i$-th stage is computed as
\begin{equation}
	{{X}_{i}}'=\operatorname{GFCP}(\operatorname{LN}({{X}_{i-1}}))+{{X}_{i-1}}
\end{equation}
\begin{equation}
	{{X}_{i}}=\operatorname{DMB}(\operatorname{LN}({{X}_{i}}'))+{{X}_{i}}'
\end{equation}
where ${{X}_{i}}\in {{\mathbb{R}}^{(N+1)\times D}}$ is the output of the $i$-th stage and $LN$ denotes the Layer Normalization.

After $T$ stages, the encoder output is passed to a task-specific head. For classification, the final representation concatenates the classification token and the average-pooled point features. The proposed path inside DMB has two stages: feature resampling at fixed sequence slots, followed by geometry-aware state propagation.

\subsection{Feature Resampling at Sequence }
This stage adapts spatial and sequence-domain feature neighborhoods while retaining the base sequence slots. It is implemented in a dynamic SSM path (D-SSM) inside the Dynamic Mamba Block. As illustrated in Fig.~\ref{overview}(b), the block also retains the forward SSM path from PointMamba \cite{PointMamba} and the channel-flipped path from Mamba3D \cite{Mamba3D}; these standard paths provide sequence views that are not restricted by the local resampling window. We replace the original 1D causal convolution \cite{TCN} with a depthwise convolution, following prior work \cite{mambaout, GridMamba, defmamba}.
\paragraph{Predicting sampling offsets}
Offset prediction requires a local cue that describes how the center feature differs from its neighbors. We obtain this cue with local context feature aggregation (LCFA), which summarizes center-to-neighbor feature differences before the offsets are predicted, as illustrated in Fig.~\ref{overview}(d).
The CLS token bypasses sampling processing and is reattached before the SSM.
Formally, given a center point $p_i\in P$ and its associated feature $f_i \in \mathbb{R}^D$, a ball query \cite{PointNet2} identifies its local neighborhood $\mathcal{N}(i)={p_j \mid |p_j-p_i|\le r}$.
The neighboring coordinates and features are denoted by $P_{\mathcal{N}(i)} \in \mathbb{R}^{K_r \times 3}$ and $F_{\mathcal{N}(i)} \in \mathbb{R}^{K_r \times D}$, respectively, where $K_r$ is the number of neighbors.
For the central feature $f_i$, the relative feature discrepancy to a neighbor $j \in \mathcal{N}(i)$ is defined as $d_{ij}=f_j-f_i$.
Subsequently, we derive local feature statistics, namely the mean $\mu_m$ and dispersion $\mu_d$, as follows:
\begin{equation}
	{{\mu }_{m}}=\frac{1}{{{K}_{r}}}\sum\nolimits_{j\in \mathcal{N}(i)}{ {{d}_{ij}} }, \quad {{\mu }_{d}}=\frac{1}{{{K}_{r}}}\sum\nolimits_{j\in \mathcal{N}(i)}{\left| {{d}_{ij}} \right|}
\end{equation}

We project the local statistics through a low-rank layer $\phi$, producing $f_{sem}=\phi(|\mu_m|+\mu_d)$. We then retain the center information by combining this cue with the original feature:
\begin{equation}
	f_{agg} = \operatorname{Concat} \left( \mu_m + \mu_d + \alpha f_i; \,\, f_{sem} \right)
\end{equation}
where $f_{agg}$ denotes the aggregated feature for the center point, and stacking all $f_{agg}$ yields the feature map $F_{agg}\in\mathbb{R}^{N\times D}$. Here, $\alpha$ is a scaling coefficient used to preserve the contribution of the center feature.
The resulting $F_{agg}$ contains both neighborhood variation and the center feature.

An offset predictor (OffsetNet) maps this representation to a three-dimensional spatial offset and a one-dimensional sequence offset for each token. Following DAT \cite{dat}, it uses a large-kernel depthwise separable convolution \cite{Xception}, channel attention (CA) \cite{senet, defmamba}, ReLU, and a $1\times1$ convolution projection:

\begin{equation}
	\overline{{\mathrm O}}_{offset}=\mathrm{Con{{v}_{1\times 1}}}(\operatorname{ReLU}(\operatorname{CA}(\mathrm{DWConv}({{F}_{agg}}))))=[\overline{\Delta p};\overline{\Delta t}]
\end{equation}

The offsets define spatial sampling anchors and sequence-domain sampling centers. In both domains, DM3D gives larger weights to candidates nearer the predicted center and decreases the weights smoothly with spatial or index distance. We use the same Gaussian form, with different distance measures and neighborhoods:
\begin{equation}
	\label{eq9}
	\mathcal{W}(d;\sigma )=\exp (-\frac{{{d}^{2}}}{2\sigma^2 })
\end{equation}
where $d$ denotes either coordinate distance or index distance, and $\sigma$ is the corresponding learnable scale parameter in each domain.

\paragraph{Resampling in the spatial domain}
In the spatial domain, the sampling anchor is moved relative to the candidate neighborhood obtained through the initial ball query, which is referred to as dynamic spatial resampling (DSR).
It reuses $\mathcal{N}(i)$ instead of performing another neighbor search, while the learned anchor changes the Gaussian weights assigned to these candidates. For each deformed anchor $p'_i$, it computes
\begin{equation}
	p'_{i}={{p}_{i}}+\Delta {{p}_{i}}
\end{equation}
\begin{equation}
	f_i^{\prime (s)}
	=
	\sum_{j\in\mathcal{N}(i)}
	\frac{
		\mathcal{W}\!\left(\|p'_i-p_j\|_2;\sigma_s\right)
	}{
		\sum_{l\in\mathcal{N}(i)}
		\mathcal{W}\!\left(\|p'_i-p_l\|_2;\sigma_s\right)+\varepsilon
	}
	\, f_j
\end{equation}
where the spatial offset is $\Delta p=\overline{\Delta p}$, $p_j$ and $f_j$ are the coordinates and feature of candidate $j$, $f_i^{\prime(s)}$ is the resampled feature, $\sigma_s$ controls the spatial kernel width, and $\varepsilon$ prevents division by zero.

Unlike grid-based deformable operators \cite{dat, defmamba}, this operation computes its weights directly from point coordinates and therefore does not require a grid-specific relative-position bias.

\paragraph{Resampling in the sequence domain}
Offset-Guided Feature Resampling (OFR) then adjusts the feature read by each base sequence slot without changing token order. Rather than approximating discrete sorting \cite{defmamba, sodeep}, it computes the feature at slot $i$ as a Gaussian-weighted mixture of nearby source slots. As illustrated in Fig.~\ref{overview}(e), the predicted offset shifts only the center of a bounded local sampling kernel, so the output remains attached to the original slot.

To keep the sequence-domain sampling center inside its local candidate span, we use an $\Omega(i)$, which denotes the local candidate index window centered at the base index $I_i$, containing at most odd $K_t$ source tokens, and define $h_t=(K_t-1)/2$. Let $h_i^+=\max_{j\in\Omega(i)}I_j-I_i$ and $h_i^-=I_i-\min_{j\in\Omega(i)}I_j$ denote the available right and left radii of the possibly truncated window. We set
\begin{equation}
	h_i(\overline{\Delta t}_i)=
	\begin{cases}
		h_i^+, & \overline{\Delta t}_i\geq 0 \\
		h_i^-, & \overline{\Delta t}_i<0
	\end{cases}
	\qquad
	\Delta t_i=h_i(\overline{\Delta t}_i)\tanh(\overline{\Delta t}_i)
	\label{eq:bounded_dt}
\end{equation}
For an interior slot, $h_i^+=h_i^-=h_t$ and Eq.~\eqref{eq:bounded_dt} reduces to $\Delta t_i=h_t\tanh(\overline{\Delta t}_i)$. Multiplication by the half-width, rather than by $K_t$, ensures that $s_i$ remains inside the actual candidate span. Here, $\Delta p\in\mathbb{R}^{N\times3}$ deforms spatial anchors, while $\Delta t\in\mathbb{R}^{N\times1}$ shifts only the center of the local sequence-domain sampling kernel.

Specifically, Eq.~\eqref{eq:bounded_dt} maps the raw sequential offset to a continuous sampling center inside $\Omega(i)$. Let $\tilde f_j^{(s)}$ denote the gated spatially resampled feature at source slot $j$. The sequence-domain feature at output slot $i$ is
\begin{equation}
	s_i=I_i+\Delta t_i
\end{equation}
\begin{equation}
	W_{ij}=\frac{\mathcal{W}(s_i-I_j;\sigma_t)}
	{\sum_{l\in\Omega(i)}\mathcal{W}(s_i-I_l;\sigma_t)},
	\quad j\in\Omega(i)
\end{equation}
\begin{equation}
	f_i^{\prime(t)}=\sum_{j\in\Omega(i)} W_{ij}\tilde f_j^{(s)}
\end{equation}
Here, $f_i^{\prime(t)}$ is the resampled feature at output slot $i$, $\sigma_t$ controls the smoothness of the local kernel, and $\tilde f_j^{(s)}=f_j^{\prime(s)}+g_j f_j$, with $g_j=\operatorname{sigmoid}(\|f_j\|_2)$, fuses the spatially resampled and original features. The normalization is row-wise over $\Omega(i)$; consequently, OFR is neither a doubly stochastic nor a one-to-one assignment matrix.

$\sigma_t$ is constrained to be positive via softplus. For $\sigma_t>0$, the Gaussian weights are continuous and differentiable with respect to $s_i$. The more detailed derivation of the limiting cases is provided in the Appendix:

\begin{itemize}
	\item As $\sigma_t \to 0^{+}$, the feature at slot $i$ approaches that of its nearest local source candidate when the nearest candidate is unique; tied nearest candidates receive equal limiting weight.
	\item For any fixed $\sigma_t>0$, the normalized Gaussian weights remain continuous and differentiable with respect to $s_i$, enabling gradient propagation to the sequence offsets.
	\item As ${{\sigma }_{t}}\to +\infty $, OFR degenerates into local average pooling over the candidate window, with gradients vanishing as the derivative converges to zero everywhere.
\end{itemize}

As $\sigma_t$ decreases, OFR approaches local nearest-source sampling away from ties; as it increases, OFR approaches local average pooling. The spatial and sequence-domain operations together form the feature-resampling stage: one changes the geometric sampling anchor, and the other changes the support read by each fixed sequence slot.

\subsection{Geometry-Aware State Propagation}
\label{sec:state_update}
Feature resampling changes the information read at fixed sequence slots but cannot eliminate every geometric discontinuity between consecutive slots. Standard Mamba predicts the step size $\Delta$ from feature cues alone and therefore does not explicitly account for the distance between consecutive deformed anchors. We implemented a continuity-aware state update (CASU), which uses that distance to modulate how much historical state is retained.

We compute the Euclidean distance between consecutive deformed anchors $p'_{i-1}$ and $p'_i$ at adjacent base sequence slots, and use it as a geometry cue.
This distance is used to define a scaling factor $\phi$ that modulates the original step size $\Delta_i$, yielding the geometry-aware step $\Delta'_i$:
\begin{equation}
	\Delta'_i = \Delta_i \cdot \phi = \Delta_i \cdot \left( 1 + \tanh \left( \| p'_i - p'_{i-1} \|_2 \right) \right)
\end{equation}

The geometric factor is applied after Softplus, because $\Delta_i>0$ and $\phi\in[1,2)$, the modulated step remains positive, whereas applying the factor before Softplus would reduce the output for negative pre-activations and reverse the intended relationship between spatial distance and state decay. As the distance grows, $\phi$ and hence $\Delta'_i$ increase. Under the stable state parameterization used by Mamba, this strengthens the decay of the previous state through $\bar{A}$ and relatively increases the contribution of the current input through $\bar{B}$, reducing historical-state propagation across larger spatial gaps. CASU is applied only to D-SSM; the parallel F-SSM and C-SSM branches retain their original updates and provide complementary sequence context. The Appendix reports the convergence of a D-SSM-only configuration.

\paragraph{Integrating the three paths}
F-SSM, C-SSM, and D-SSM provide different sequence views. To obtain the block output, we use a lightweight cross-path interaction before combining them.

The three branch features, $F_F, F_C, F_D \in \mathbb{R}^{N \times D}$, are first adaptively modulated by the other two branches before fusion:
\begin{equation}
	w_i = \operatorname{sigmoid}(F_i), \quad F'_i = F_i \odot \frac{1}{2}(w_j + w_k)
\end{equation}
where $(i,j,k)$ cycles over $(F,C,D)$, $(C,D,F)$, and $(D,F,C)$.


\begin{table*}[!ht]
	\centering
	\footnotesize
	\caption{ Classification on ModelNet40 and ScanObjectNN. We report overall accuracy (\%), number of parameters (\#P), and FLOPs (\#F). We use \textit{rotation} and \textit{scale\&translate} as data augmentation for ScanObjectNN and ModelNet40, respectively. Values marked with $\pm$ are means and standard deviations over three runs.
	}
	\begin{tabular}{rlccccccc}
		\toprule
		\multirow{2}{*}{ Reference}         & \multirow{2}{*}{Methods}              & \multicolumn{3}{c}{ScanObjectNN} & \multicolumn{2}{c}{ModelNet40 (1k pts)} & \multirow{2}{*}{\#P (M)}    & \multirow{2}{*}{\#F (G)}                               \\
		\cmidrule(l){3-5} \cmidrule(l){6-7} &                                       & OBJ\_BG                          & OBJ\_ONLY                               & PB\_T50\_RS                 & w/o Vote                 & w/ Vote       & ~    & ~    \\ \toprule
		\multicolumn{9}{c}{\textit{Supervised Learning Only}}                                                                                                                                                                                           \\ \hline
		CVPR 17                             & PointNet \cite{pointnet}              & 73.3                             & 79.2                                    & 68.0                        & 89.2                     & -             & 3.5  & 0.5  \\
		NeurIPS 17                          & PointNet++   \cite{PointNet2}         & 82.3                             & 84.3                                    & 77.9                        & 90.7                     & -             & 1.5  & 1.7  \\
		TOG 19                              & DGCNN  \cite{dgcnn}                   & 82.8                             & 86.2                                    & 78.1                        & 92.9                     & -             & 1.8  & 2.4  \\
		NeurIPS 22                          & PointNeXt \cite{PointNeXt}            & -                                & -                                       & 87.7                        & 92.9                     & -             & 1.4  & 3.6  \\
		JAS 23                              & PointConT  \cite{pointcont}           & -                                & -                                       & 88.0                        & 93.5                     & -             & -    & -    \\
		NeurIPS 24                          & PointMamba \cite{PointMamba}          & 88.30                            & 87.78                                   & 82.48                       & -                        & -             & 12.3 & 3.6  \\
		ACM MM 25                           & HydraMamba  \cite{HydraMamba}         & -                                & -                                       & 88.3                        & \textbf{94.0}            & -             & -    & -    \\

		AAAI   25                           & PCM  \cite{PCM}                       & -                                & -                                       & 88.1                        & 93.4                     & -             & 34.2 & 45.0 \\
		KBS   25                            & LFE-PointMamba  \cite{LFE-PointMamba} & 92.3                             & 90.6                                    & 88.5                        & 93.0                     & -             & 9    & 2.8  \\
		CVPR   25                           & SAST  \cite{SAST}                     & 92.25                            & 91.39                                   & 87.30                       & 92.7                     & -             & 12.3 & 3.6  \\
		~                                   & DM3D                                  & \textbf{93.11}                   & \textbf{91.74}                          & \textbf{90.83}              & \textbf{94.0}            & -             & 18.6 & 4.0  \\ \hline
		\multicolumn{9}{c}{\textit{With Self-supervised Pre-training}}                                                                                                                                                                                  \\ \hline
		CVPR 22                             & Point-BERT \cite{Point-BERT}          & 87.43                            & 88.12                                   & 83.07                       & 92.7                     & 93.2          & 23.8 & 4.8  \\
		ECCV 22                             & Point-MAE \cite{point-mae}            & 92.77                            & 91.22                                   & 89.04                       & 92.7                     & 93.8          & 23.8 & 4.8  \\
		NeurIPS 23                          & PointGPT-S   \cite{Pointgpt}          & 93.39                            & 92.43                                   & 89.17                       & 93.3                     & 94.0          & 29.2 & 5.7  \\
		AAAI 24                             & Point-FEMAE \cite{point-femae}        & \textbf{95.18}                   & 93.29                                   & 90.22                       & 94.0                     & 94.5          & 27.4 & 3.6  \\
		NeurIPS 24                          & PointMamba  \cite{PointMamba}         & 94.32                            & 92.60                                   & 89.31                       & 93.6                     & 94.1          & 12.3 & 3.6  \\
		ACM MM 24                           & Mamba3D \cite{Mamba3D}                & 93.12                            & 92.08                                   & 92.05                       & 94.7                     & 95.1          & 16.9 & 3.9  \\
		CVPR  25                            & SAST   \cite{SAST}                    & 94.32                            & 91.91                                   & 89.10                       & 93.4                     & -             & 12.3 & 3.6  \\
		ICCV   25                           & StruMamba3D  \cite{strumamba3d}       & 95.18                            & 93.63                                   & 92.75                       & \textbf{95.1}            & \textbf{95.4} & 15.8 & 4.0  \\
		KBS   25                            & LFE-PointMamba  \cite{LFE-PointMamba} & 92.86                            & 93.37                                   & 89.32                       & -                        & -             & 9    & 2.8  \\
		ICCV 25                             & Point-PQAE \cite{point-pqae}          & 95.0                             & 93.6                                    & 89.6                        & 94.0                     & -             & 22.1 & -    \\
		~                                   & DM3D                                  & 94.71$_{\pm 0.29}$               & \textbf{93.83$_{\pm 0.37}$}             & \textbf{93.30$_{\pm 0.41}$} & 94.9$_{\pm 0.23}$        & 95.2          & 18.6 & 4.0  \\ \bottomrule
	\end{tabular}
	\label{Class}
\end{table*}

Each branch is thus modulated by the activation strength of the other two. The modulated features are concatenated and projected back to $D$ channels by a grouped $1\times1$ convolution:
\begin{equation}
	F_{\mathrm{three}}=\operatorname{GConv}_{1\times1}
	\left(\operatorname{Concat}\left(F'_F,F'_C,F'_D\right)\right),
\end{equation}
where $\operatorname{GConv}$ is the grouped convolution layer and $F_{\mathrm{three}}$ is the output of DMB.

\begin{table}[!t]
	\centering
	\caption{Implementation details and hyperparameter settings. S\&T denotes Scale\&Translation.}
	\footnotesize
	\begin{tabular}{lccc}
		\toprule
		Config                  & Pre-training      & Classification    & Part Seg.         \\ \midrule
		Setting                 & ShapeNet          & MN40/ScanNN       & ShapeNetPart      \\
		Optimizer               & AdamW             & AdamW             & AdamW             \\
		Learning rate           & $1\mathrm{e}{-3}$ & $5\mathrm{e}{-4}$ & $2\mathrm{e}{-4}$ \\
		Weight decay            & 0.05              & 0.05              & 0.05              \\
		Scheduler               & CosLR             & CosLR             & CosLR             \\
		Warmup epochs           & 10                & 10                & 10                \\
		Drop path rate          & 0.1               & 0.1               & 0.3               \\
		Training epochs         & 300               & 300               & 300               \\
		Batch size              & 64                & 32                & 32                \\ \hline
		Input points            & 1024              & 1024/2048         & 2048              \\
		Hidden dim. ($D$)       & 384               & 384               & 384               \\
		Stage layers ($T$)      & 12                & 12                & 12                \\
		FPS samples ($N$)       & 64                & 128               & 128               \\
		KNN neighbors ($K$)     & 32                & 32                & 16                \\
		Ball query radius ($r$) & 0.25              & 0.2/0.25          & 0.25              \\
		OffsetNet kernel size   & 5                 & 5                 & 5                 \\
		DSR $K_r$               & 4                 & 4                 & 4                 \\
		OFR $K_t$               & 9                 & 9                 & 9                 \\
		DSR $\sigma_s$          & 1                 & 1                 & 1                 \\
		OFR $\sigma_t$          & 0.2               & 0.2               & 0.2               \\
		Augmentation            & Rotation          & S\&T / Rotation   & Scale\&Center     \\ \bottomrule
	\end{tabular}
	\label{tab:settings}
\end{table}

\section{Experiments}
\subsection{Implementation Details}
All experiments are conducted on a single NVIDIA TITAN RTX 24GB GPU.
We set the total number of stages $T$ to 12, the hidden dimension $D$ to 384, the number of points sampled by FPS to $N=128$, and the number of neighbors in KNN to $K=32$.
For DSR and OFR, $K_r$ and $K_t$ are set to 4 and 9, while $\sigma_s$ and the directly learned $\sigma_t$ are initialized to 1 and 0.2, respectively. The learned $\sigma_t$ remains positive in the empirical observation.
The detailed configurations are given in Table~\ref{tab:settings}. The pre-training follows Point-MAE protocols \cite{point-mae, PointMamba, Mamba3D} on ShapeNetCore \cite{ShapeNet}; classification and part segmentation use cross-entropy loss.

\subsection{Downstream Tasks}
\paragraph{Classificatory identification}
We use a three-layer MLP as the classification head and evaluate our DM3D on ModelNet40 \cite{modelnet} and ScanObjectNN \cite{ScanObjectNN}.
ModelNet40 is a synthetic CAD benchmark and uses 1024 input points with scale\&translate augmentation. ScanObjectNN is a real-world scanned-object dataset with noise, partial observations, and occlusions; it uses 2048 input points with rotation augmentation.
Table~\ref{Class} compares DM3D with supervised and pretrained point cloud models. Without pre-training, DM3D achieves 94.0\% on ModelNet40 and 90.83\% on PB\_T50\_RS, outperforming the listed recent strong Mamba variants, including SAST \cite{SAST} and PCM \cite{PCM}, HydraMamba \cite{HydraMamba}.

With pre-training, DM3D further improves PB\_T50\_RS overall accuracy (OA) to $93.30\pm0.41\%$, surpassing recent counterparts including StruMamba3D \cite{strumamba3d}, Point-PQAE \cite{point-pqae} (Transformer-based, 89.6\%), and SAST \cite{SAST} (89.1\%). Across three runs, the standard deviations of the reported non-voting results range from 0.23 to 0.41 percentage points, indicating performance is relatively stable under the protocol.
The gain is most pronounced on PB\_T50\_RS. Together with the improvements over the corresponding PointMamba \cite{PointMamba} and LFE-PointMamba \cite{LFE-PointMamba} models, this result indicates that changing the features aggregated within a fixed sequence can improve recognition when the original local context is unreliable.
DM3D also uses fewer parameters than PCM \cite{PCM} and Point-PQAE \cite{point-pqae} in this table, while adding 0.1G FLOPs relative to Mamba3D \cite{Mamba3D}.

\begin{figure*}[!t]
	\centering
	\includegraphics[width=0.85\linewidth]{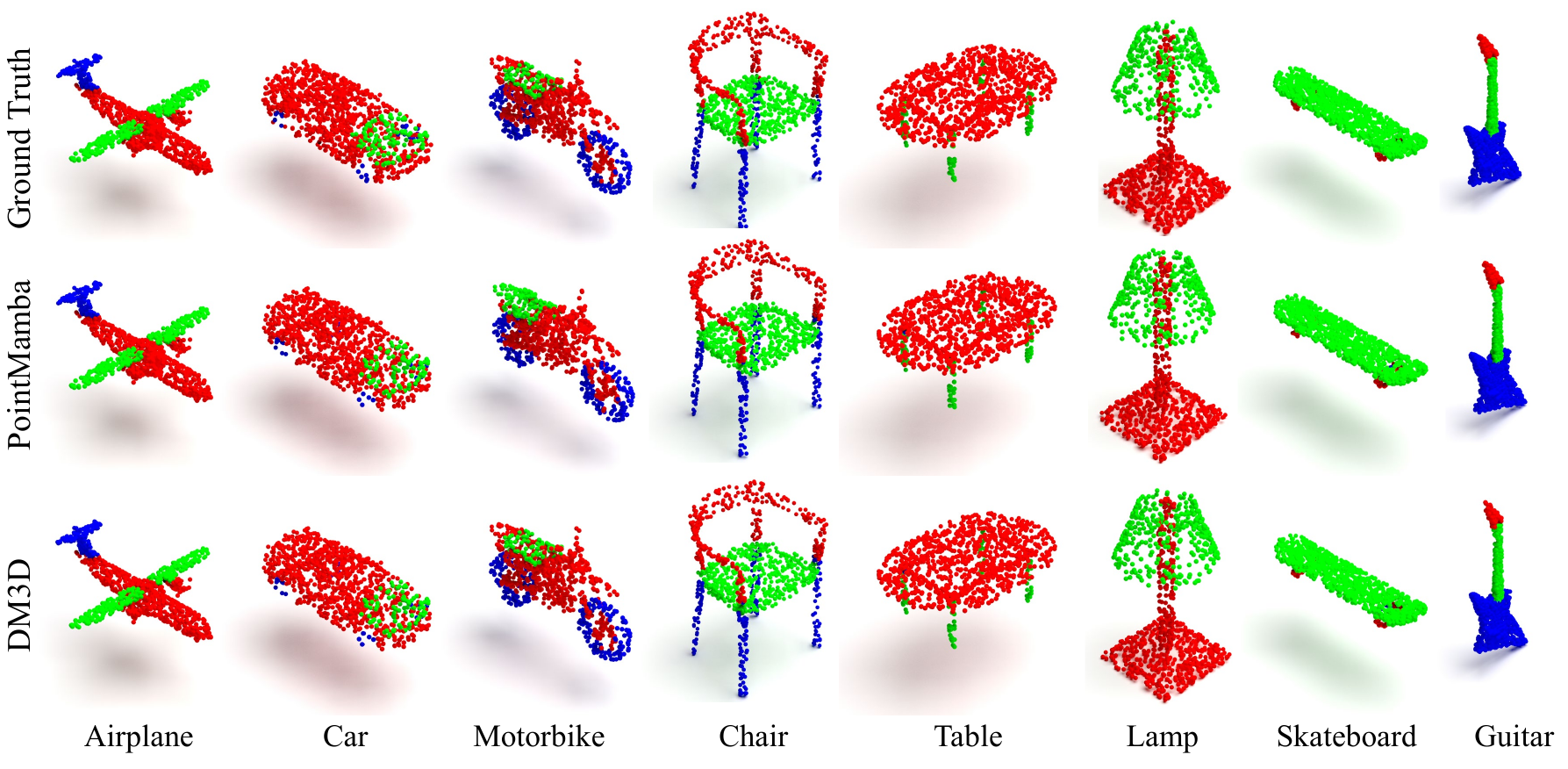}
	\caption{Qualitative part segmentation results on ShapeNetPart.}
	\label{fig:partseg}
\end{figure*}

\begin{table}[!t]
	\centering
	\footnotesize
	\caption{{Few-shot classification on ModelNetFewShot}. A$_{\pm\text{std}}$ represents the average (A) and standard deviation (std) of OA.
	}
	\begin{tabular}{lcccc}
		\toprule
		\multirow{2}{*}{Methods}     & \multicolumn{2}{c} {5-Way} & \multicolumn{2}{c}{10-Way}                                                          \\
		\cmidrule
		(l){2-3} \cmidrule(l){4-5}
		~                            & 10-Shot                    & 20-Shot                    & 10-Shot                   & 20-Shot                    \\ \toprule
		\multicolumn{5}{c}{\textit{Supervised Learning Only}}                                                                                           \\ \hline
		PointNet \cite{pointnet}     & 52.0$_{\pm3.8}$            & 57.8$_{ \pm4.9}$           & 46.6$_{ \pm4.3}$          & 35.2$_{ \pm4.8}$           \\
		DGCNN \cite{dgcnn}           & 31.6$_{\pm2.8}$            & 40.8$_{ \pm4.6}$           & 19.9$_{ \pm2.1}$          & 16.9$_{ \pm1.5}$           \\
		OcCo \cite{OcCo}             & 90.6$_{ \pm2.8}$           & 92.5$_{ \pm1.9}$           & 82.9$_{ \pm1.3}$          & 86.5$_{ \pm2.2}$           \\
		DM3D                         & \textbf{91.1$_{ \pm5.2}$}  & \textbf{95.0$_{ \pm2.9}$}  & \textbf{86.1$_{ \pm4.6}$} & \textbf{92.0$_{ \pm1.8}$}  \\ \hline
		\multicolumn{5}{c}{\textit{With Self-supervised Pre-training}}                                                                                  \\ \hline
		ACT \cite{ACT}               & 96.8$_{ \pm2.3}$           & 98.0$_{ \pm1.4}$           & 93.3$_{ \pm4.0}$          & 95.6$_{ \pm2.8 }$          \\
		Point-BERT \cite{Point-BERT} & 94.6$_{\pm 3.1}$           & 96.3$_{\pm 2.7}$           & 91.0$_{\pm 5.4}$          & 92.7$_{ \pm5.1}$           \\
		Point-MAE \cite{point-mae}   & 96.3$_{ \pm2.5}$           & 97.8$_{ \pm1.8}$           & \textbf{92.6$_{\pm4.1}$}  & 95.0$_{ \pm3.0}$           \\
		PointGPT-S \cite{Pointgpt}   & \textbf{96.8$_{\pm 2.0}$ } & \textbf{98.6$_{ \pm1.1}$}  & 92.6$_{ \pm4.6}$          & 95.2$_{\pm 3.4}$           \\
		PointMamba \cite{PointMamba} & 95.0$_{\pm 2.3}$           & 97.3$_{\pm 1.8}$           & 91.4$_{\pm 4.4}$          & 92.8$_{ \pm4.0}$           \\
		DM3D                         & 96.1$_{ \pm4.5}$           & 97.9$_{ \pm2.3}$           & 91.4$_{\pm 4.3}$          & \textbf{95.8$_{ \pm3.0}$ } \\ \bottomrule
	\end{tabular}
	\label{fewshot}
\end{table}

\paragraph{Few-shot learning}
Following the standard N-way K-shot protocol and prior work \cite{PointMamba, Mamba3D, Point-BERT, Hymamba}, we evaluate the few-shot learning capability of DM3D on the ModelNetFewShot benchmark.
As shown in Table~\ref{fewshot}, DM3D performs best among the compared methods in all four supervised-only settings.
With pre-training, DM3D remains competitive but does not achieve the best results under the 10-shot settings, suggesting that its pretrained features require slightly more labeled data to adapt to downstream classification. As the number of labeled examples increases to 20 shots, DM3D benefits more clearly from the additional supervision and achieves its strongest relative result in the 10-way 20-shot setting. These results indicate that the pretrained DM3D representation transfers effectively to classification tasks with limited data.

\begin{table}[!t]
	\centering
	\footnotesize
	\caption{Part segmentation on ShapeNetPart. We report class-level mIoU (mIoU$_C$) and instance-level mIoU (mIoU$_I$).}
	\begin{tabular}{rlcc}
		\toprule
		Reference  & Method                               & mIoU$_C$(\%)   & mIoU$_I$(\%)   \\ \toprule
		\multicolumn{4}{c}{\textit{Supervised Learning Only}}                               \\ \hline
		CVPR 17    & PointNet \cite{pointnet}             & 80.4           & 83.7           \\
		NeurIPS 17 & PointNet++ \cite{PointNet2}          & 81.9           & 85.1           \\
		TOG 19     & DGCNN \cite{dgcnn}                   & 82.3           & 85.2           \\
		           & DM3D                                 & \textbf{ 83.7} & \textbf{ 85.3} \\ \hline
		\multicolumn{4}{c}{\textit{With Self-supervised Pre-training}}                      \\ \hline
		ECCV 22    & Point-MAE \cite{point-mae}           & 84.2           & 86.1           \\
		CVPR 22    & Point-BERT \cite{Point-BERT}         & 84.1           & 85.6           \\
		NeurIPS 23 & PointGPT-S \cite{Pointgpt}           & 84.1           & 86.2           \\
		ICLR 23    & ACT       \cite{ACT}                 & 84.7           & 86.2           \\
		NeurIPS 24 & PointMamba  \cite{PointMamba}        & 84.4           & 86.0           \\
		ACM MM 24  & Mamba3D   \cite{Mamba3D}             & 83.6           & 85.6           \\
		ICCV 25    & Point-PQAE  \cite{point-pqae}        & 84.6           & 86.1           \\
		ICCV 25    & StruMamba3D      \cite{strumamba3d}  & -              & 86.7           \\
		KBS 25     & LFE-PointMamba \cite{LFE-PointMamba} & 84.3           & 86.1           \\
		AAAI 26    & CloudMamba      \cite{CloudMamba}    & -              & 86.6           \\
		           & DM3D                                 & \textbf{84.8}  & \textbf{86.7}  \\ \bottomrule
	\end{tabular}
	\label{partseg}
\end{table}

\paragraph{Part segmentation}
ShapeNetPart \cite{ShapeNet} contains 50 part categories across 16 object classes.
We use 2048 input points without normals, and adopt a PointNet++-style \cite{PointNet2, PointMamba} segmentation head  that aggregates features from the 4th, 8th, and 12th encoder layers.
With pre-training, DM3D achieves 84.8\% mIoU$_C$ and 86.7\% mIoU$_I$, as reported in Table~\ref{partseg}. It outperforms PointMamba by 0.7 and 0.4 percentage points in mIoU$_I$ and mIoU$_C$, respectively, and achieves comparable or slightly better performance than recent methods, including Point-PQAE \cite{point-pqae}, StruMamba3D \cite{strumamba3d}, LFE-PointMamba \cite{LFE-PointMamba} and CloudMamba \cite{CloudMamba}. Overall, the results demonstrate the effectiveness of DM3D for fine-grained point cloud part segmentation. Fig.~\ref{fig:partseg} provides qualitative comparisons with PointMamba and the ground truth.

Fig.~\ref{fig:partseg} presents qualitative segmentation results for eight representative ShapeNetPart categories. Both PointMamba and DM3D capture the major semantic parts and produce predictions that are largely consistent with the ground-truth annotations. In “Motorbike” cases, DM3D exhibits slightly cleaner local predictions, although the overall differences between the two methods are relatively subtle. See Appendix for more detailed visual comparison.

\begin{figure}[!t]
	\centering
	\includegraphics[width=3in]{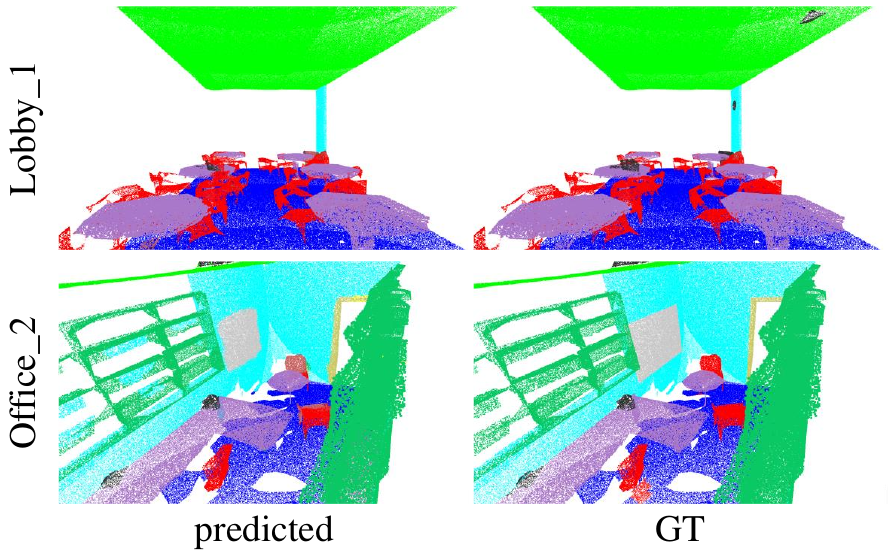}
	\caption{Semantic segmentation on the real-world S3DIS dataset.  }
	\label{s3dis}
\end{figure}

\paragraph{Scene-level semantic segmentation}
We additionally evaluate DM3D on S3DI \cite{s3dis}, a real-world indoor-scene dataset, using the Area~5 split. We train on Areas 1--4 and 6 and test on Area~5. The scene configuration uses a hidden dimension of 96, a drop-path rate of 0.1, AdamW with an initial learning rate of $1\mathrm{e}{-4}$ and weight decay of 0.05, five warmup epochs, a batch size of four, and 128 training epochs.
PTv3 and CloudMamba is not included in the S3DIS benchmark as its data split, input resolution, and other evaluation protocols differ from ours, precluding direct comparison.
As shown in Table~\ref{semseg}, DM3D achieves 71.0\% mAcc and 65.3\% mIoU, outperforming the listed comparison methods in mIoU. The PCM result follows the public implementation reported in \cite{CloudMamba}. This result demonstrates the applicability of DM3D to real-world indoor scene scans (see Fig.~\ref{s3dis} for visualization).

\begin{table}[!t]
	\centering
	\footnotesize
	\caption{Scene-level semantic segmentation on S3DIS Area~5. We report mean accuracy (mAcc) and mean IoU (mIoU).}
	\begin{tabular}{rlcc}
		\toprule
		Reference  & Method                       & mAcc (\%) & mIoU (\%) \\ \midrule
		NeurIPS 17 & PointNet++ \cite{PointNet2}  & 67.1      & 53.5      \\
		CVPR 22    & Point-BERT \cite{Point-BERT} & 70.3      & 60.8      \\
		ECCV 22    & Point-MAE \cite{point-mae}   & 69.9      & 60.8      \\
		NeurIPS 23 & PointGPT-L \cite{Pointgpt}   & 70.6      & 62.2      \\
		ICLR 23    & ACT \cite{ACT}               & 71.1      & 61.2      \\
		ICCV 25    & Point-PQAE \cite{point-pqae} & 70.6      & 61.4      \\
		AAAI 25    & PCM \cite{PCM}               & -         & 63.4      \\
		           & DM3D                         & 71.0      & 65.3      \\ \bottomrule
	\end{tabular}
	\label{semseg}
\end{table}

\paragraph{Efficiency analysis}
Using the notation defined in Table~\ref{tab:settings}, and $G$ as the number of groups in the grouped $1\times1$ convolution of path integration. The additional computational complexity introduced by DM3D is
\begin{equation}
	\Delta C=\mathcal{O}(BND[\!{{{K}_{r}}}\!+\!{{{K}_{t}}}\!+\!{D}\!+\!{{}^{D}\!\!\diagup\!\!{}_{G}\;}\!+\!{1}])
\end{equation}

\begin{figure}[!t]
	\centering
	\includegraphics[width=1.6in]{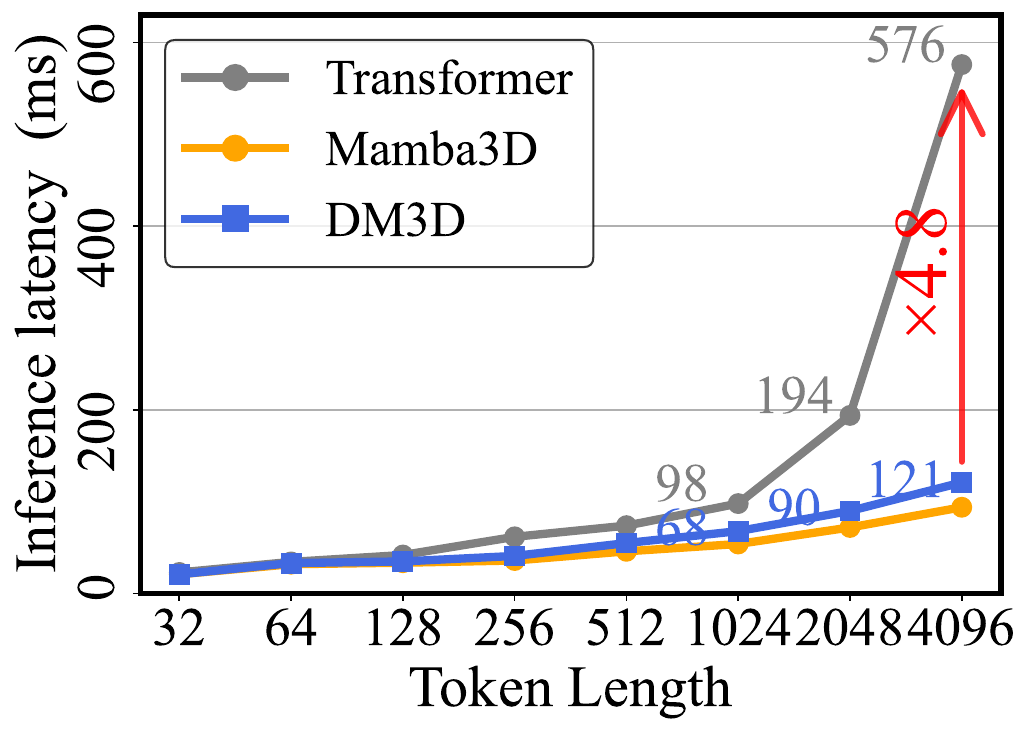}
	\hfil
	\includegraphics[width=1.6in]{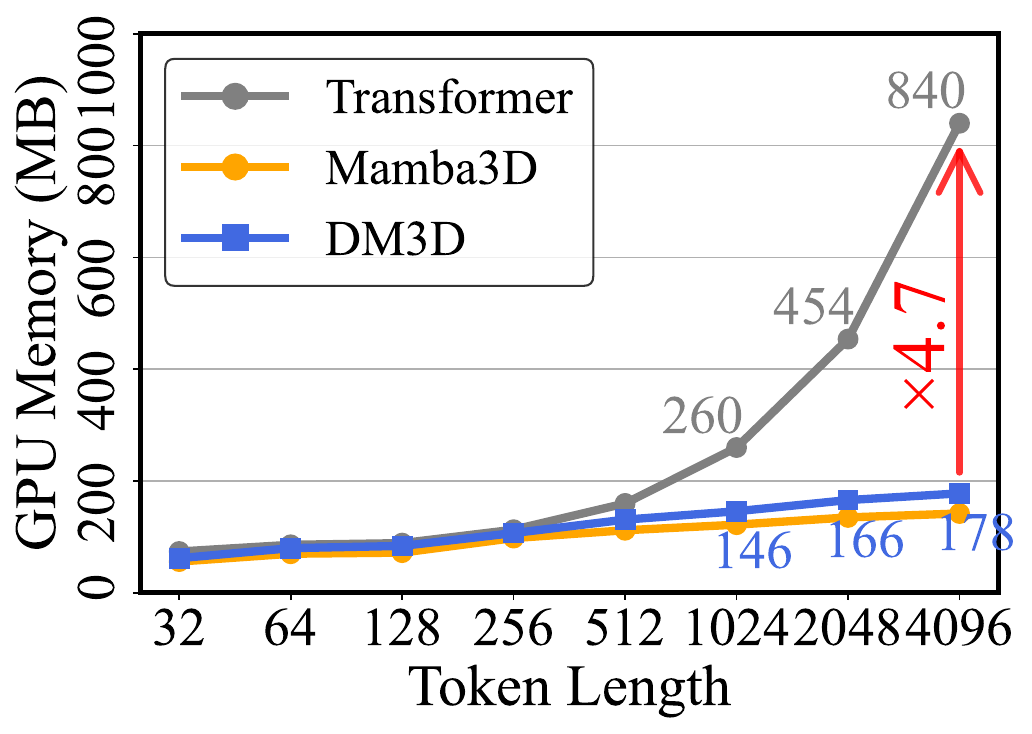}
	\caption{Measured inference latency and peak GPU memory usage for one sample.}
	\label{fig_cost}
\end{figure}

LCFA and DSR contribute $\mathcal{O}(BNDK_r)$, OFR contributes $\mathcal{O}(BNDK_t)$, OffsetNet contributes $\mathcal{O}(BND^2)$, path integration contributes $\mathcal{O}(BND^2/G)$, and CASU adds $\mathcal{O}(BND)$. Because $D$, $G$, $K_r$, and $K_t$ are fixed, the added cost grows linearly with $N$.
With fixed $D$, $G$, $K_r$, and $K_t$, the added cost grows is linear for $N$ (see Table~\ref{fig_cost}), as DM3D uses local-window operations.

We measure inference latency, peak GPU memory usage, and throughput on an NVIDIA TITAN RTX with a batch size of 1. The measurements characterize model inference. Fig.~\ref{fig_cost} consolidates the measured latency and memory comparison, while Table~\ref{tab:fps} reports the corresponding throughput. At $N=4096$, DM3D has $\sim4.8\times$ lower latency and $\sim4.7\times$ lower peak memory than the Transformer reference, and achieves 15.8 FPS vs 1.7 FPS. It slight slower than Mamba3D because DSR and OFR introduce local gather-and-weight operations. These results explain model efficiency.

\begin{table}[!t]
	\centering
	\footnotesize
	\caption{Throughput in frames per second (FPS) on a TITAN RTX with batch size one.}
	\begin{tabular}{lcccc}
		\toprule
		Token length & 128   & 1024 & 2048 & 4096 \\ \midrule
		Transformer  & 128.4 & 30.6 & 13.0 & 1.7  \\
		Mamba3D      & 141.5 & 63.7 & 35.2 & 18.6 \\
		DM3D         & 139.3 & 60.1 & 31.9 & 15.8 \\ \bottomrule
	\end{tabular}
	\label{tab:fps}
\end{table}

\subsection{Ablation Studies}
We organize the ablations around the design, all variants are trained from scratch.

\paragraph{Evidence for the dynamic path}
The local OFR window adapts the features entering D-SSM but does not replace the broader sequence context provided by the standard paths. As shown in Table~\ref{D-SSM}, D-SSM alone reaches 86.9\% on PB\_T50\_RS, improving upon F-SSM alone by 2.5\%. The two standard paths together achieve 88.7\%, indicating that their complementary sequence views remain important. Combining D-SSM with both standard paths yields the best accuracy of 90.8\%, exceeding F-SSM+C-SSM by 2.1\% and D-SSM alone by 3.9\%. These results show that the dynamic path contributes useful adaptive features while benefiting from the global sequence context retained by the standard paths.

\begin{table}[t]
	\centering
	\footnotesize
	\caption{Contribution of the dynamic and standard paths on PB\_T50\_RS.}
	\begin{tabular}{cccc}
		\toprule
		F-SSM     & C-SSM     & D-SSM     & PB\_T50\_RS \\ \midrule
		\ding{51} & \ding{55} & \ding{55} & 84.4\%      \\
		\ding{55} & \ding{55} & \ding{51} & 86.9\%      \\
		\ding{51} & \ding{51} & \ding{55} & 88.7\%      \\
		\ding{51} & \ding{51} & \ding{51} & 90.8\%      \\ \bottomrule
	\end{tabular}
	\label{D-SSM}
\end{table}

\paragraph{Evidence for the components design}
Table~\ref{Ablation model} evaluates each component by removing it from the full model. Without LCFA, accuracy decreases by 1.3\%/1.4\% on OBJ\_ONLY/PB\_T50\_RS. Removing DSR or OFR causes larger drops of 2.4\%/2.2\% and 2.6\%/3.5\%, respectively, showing that adaptation in both the spatial and sequence domains contributes to feature resampling. Disabling CASU reduces accuracy by 1.9\%/2.2\%, while removing path integration results in decreases of 0.8\%/1.8\%. The full model performs best on both splits, supporting the joint use of local offset cues, two-domain resampling, geometry-aware state propagation, and path integration.

\begin{table}[!t]
	\centering
	\footnotesize
	\caption{{Component ablation on OBJ\_ONLY and PB\_T50\_RS.} Accuracy values are overall accuracy (\%).
	}
	\begin{tabular}{lcccc}
		\toprule
		Method                              & OBJ\_ONLY     & PB\_T50\_RS   & \#P (M) & \#F (G) \\ \toprule
		$\star$ Full                        & \textbf{91.7} & \textbf{90.8} & 18.6    & 4.0     \\
		\hspace{0.5em}	w/o LCFA             & 90.4          & 89.4          & 18.6    & 4.0     \\
		\hspace{0.5em}	w/o DSR              & 89.3          & 88.6          & 18.6    & 4.0     \\
		\hspace{0.5em}	w/o OFR              & 89.1          & 87.3          & 18.6    & 4.0     \\
		\hspace{0.5em}	w/o CASU             & 89.8          & 88.6          & 18.6    & 4.0     \\
		\hspace{0.5em}	w/o Path integration & 90.9          & 89.0          & 18.5    & 4.0     \\ \bottomrule
	\end{tabular}
	\label{Ablation model}
\end{table}

\paragraph{Effect of resampling operators}
Table~\ref{ablation_param} first evaluates how the sequence offset and resampling operator affect OFR. Learned offsets achieve 91.74\%/90.80\% on OBJ\_ONLY/PB\_T50\_RS, improving over zero offsets by 2.84/4.20 percentage points; random offsets perform worse than zero offsets, indicating that the gain depends on input-adaptive displacement rather than perturbation alone. Using the full OFR result as reference, replacing OFR with no resampling, local convolution, or Sinkhorn reduces accuracy by 3.30/4.36, 2.40/3.68, and 2.15/2.88 points, respectively. These comparisons support learned, candidate-dependent resampling within the fixed local window.


\begin{table}[!t]
	\centering
	\footnotesize
	\caption{
		Ablation of the offset setting and sequence-domain resampling operator in OFR.
	}
	\begin{tabular}{lcc}
		\toprule
		Variants                  & OBJ\_ONLY      & PB\_T50\_RS    \\ \toprule
		\multicolumn{3}{c}{\textit{(a) Offset methods }}            \\ \hline

		Zero                      & 88.9           & 86.6           \\
		Random                    & 87.36          & 84.87          \\
		Learned(Ours)             & \textbf{91.74} & \textbf{90.8}  \\ 	   \hline
		\multicolumn{3}{c}{\textit{(b) Resampling methods }}        \\ \hline

		No resampling             & 88.44          & 86.47          \\
		Local convolution         & 89.34          & 87.15          \\
		Sinkhorn  \cite{Sinkhorn} & 89.59          & 87.95          \\
		OFR (Ours)                & \textbf{91.74} & \textbf{90.83} \\
		\bottomrule
	\end{tabular}
	\label{ablation_param}
\end{table}

\paragraph{Geometry-aware state modulation}
To evaluate the design of CASU, we compare it with two alternative modulation strategies: Linear Scaling, which sets $\phi = 2\|p_i-p_{i-1}\|_2$, and Feature Similarity, which sets $\phi = 1 + [1-\cos(f_i, f_{i-1})]/2$. Here, $p_{i-1}$ and $p_i$ denote consecutive deformed anchors, and $f_{i-1}$ and $f_i$ denote their corresponding features.
Linear Scaling is a simple geometry-based baseline, while Feature Similarity is a lightweight feature-based variant. They are used to test whether state updates should be guided by geometric continuity or feature cues.

As shown in Table~\ref{tab:casu_ablation}, bounded nonlinear CASU performs better than  Linear Scaling and Feature Similarity on both splits.  Linear Scaling uses the same geometric distance but scales it without an upper bound, whereas Feature Similarity replaces distance with feature similarity.  Their lower accuracies indicate that bounded distance modulation is more effective.

The selective-activation results in Table~\ref{tab:casu_ablation} examine where CASU is most effective under a fixed 30\% activation. Specifically, CASU is applied only to a selected 30\% of the adjacent-anchor gaps, while the remaining positions use the original state update without CASU.
Activating CASU at positions with the largest adjacent-anchor distances achieves 90.75\% on OBJ\_ONLY and 89.55\% on PB\_T50\_RS, outperforming random activation by 0.54\% and 0.75\%, respectively.
In contrast, activating CASU at the smallest-gap positions yields 89.83\% and 88.92\%, which are 0.92\% and 0.63\% lower than the largest-gap setting. These results indicate that CASU is more beneficial at positions associated with larger geometric discontinuities. Nevertheless, full CASU remains the strongest setting, reaching 91.74\% and 90.83\%, and exceeds largest-gap activation by 0.99 and 1.28 points. This suggests that large-gap positions contribute more strongly to the effectiveness of CASU, while state modulation at the remaining positions still provides complementary gains.

\begin{table}[!t]
	\centering
	\footnotesize
	\caption{Ablation of CASU. ``Top'' and ``Bottom'' activate CASU at positions with the largest and smallest adjacent-anchor distances, respectively.}
	\begin{tabular}{lcc}
		\toprule
		State update       & OBJ\_ONLY (\%) & PB\_T50\_RS (\%)               \\ \midrule
		\multicolumn{3}{c}{\textit{(a) State update modulation strategies.}} \\\hline
		Linear Scaling     & 90.88          & 89.02                          \\
		Feature Similarity & 90.01          & 88.23                          \\
		CASU               & \textbf{91.74} & \textbf{90.83}                 \\\hline
		\multicolumn{3}{c}{\textit{(b) Selective activation strategies.}}    \\\hline
		Random 30\%        & 90.21          & 88.80                          \\
		Top 30\%           & 90.75          & 89.55                          \\
		Bottom 30\%        & 89.83          & 88.92                          \\ \bottomrule
	\end{tabular}
	\label{tab:casu_ablation}
\end{table}

\subsection{Feature Support Analysis}

\begin{figure*}[!t]
	\centering
	\includegraphics[width=0.98\linewidth]{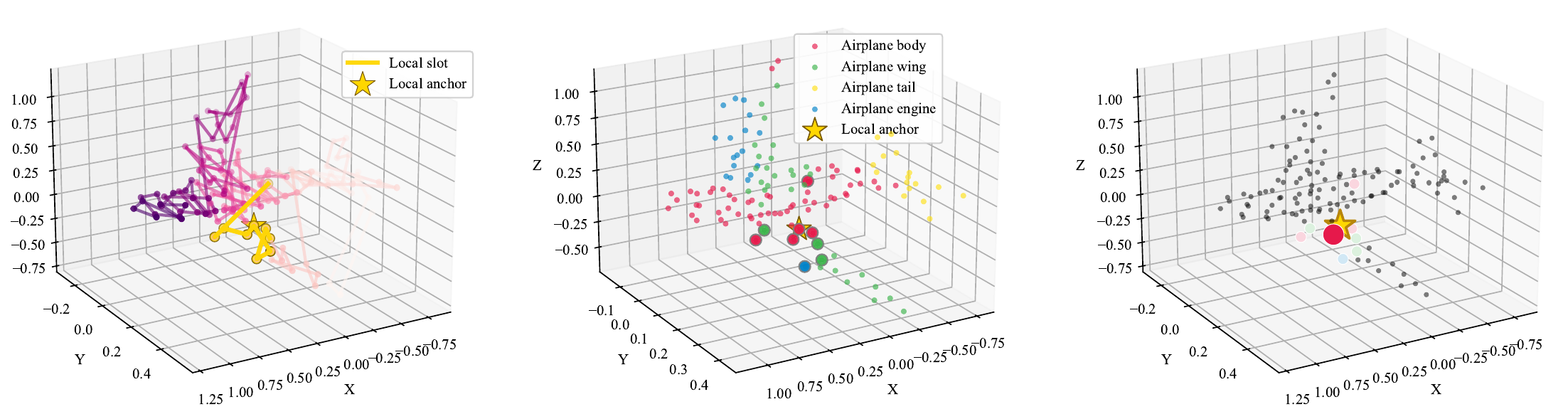}
	\includegraphics[width=0.98\linewidth]{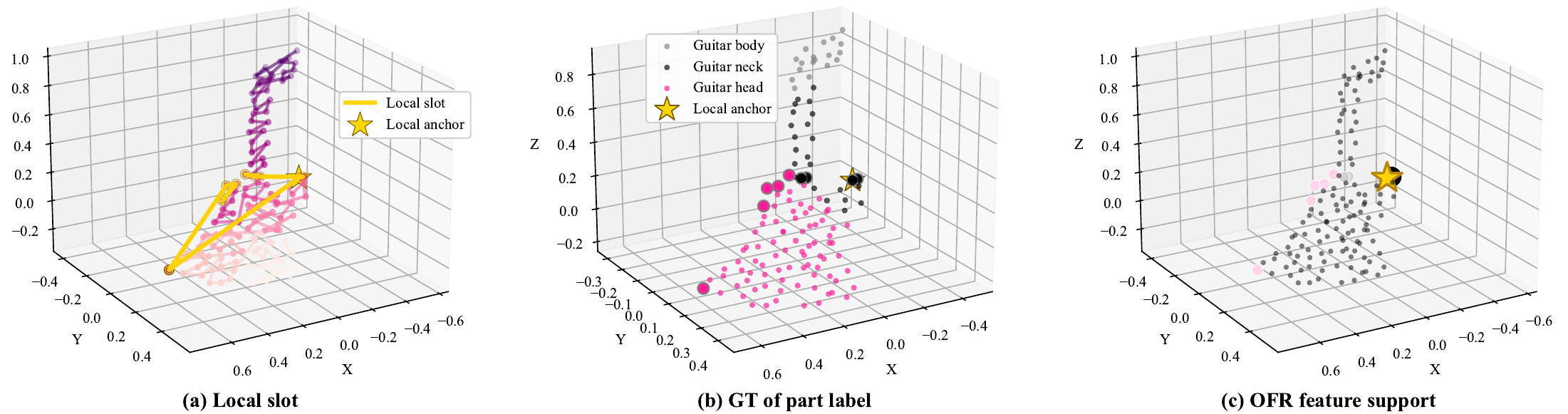}
	\caption{Local-support patterns for an Airplane (top) and a Guitar (bottom) of ShapeNetpart. {(a) Local slot:} 9-slot Hilbert windows may contain large 3D jumps. {(b) GT of part label:} The highlighted windows span multiple object parts, with the yellow star marking the query anchor. {(c) OFR feature support:} OFR concentrates support on a few spatially relevant candidates within the fixed window rather than globally reordering the sequence.}
	\label{fig:support_weights}
\end{figure*}

\paragraph{Support metrics}

To evaluate whether the sequence-domain support selected by OFR is also locally consistent in 3D space, we compute the weighted geometric distance $D_{\mathrm{geo}}$ and the weighted geometric support rate $S_{\mathrm{geo}}$. Both metrics use the OFR weights $W_{ij}^{(t)}$ defined previously. To avoid trivial self-contributions, the query slot is excluded, and the remaining weights over $\Omega^{-}(i)=\Omega(i)\setminus{i}$ are renormalized to unit sum, yielding $\bar{W}_{ij}^{(t)}$.

Let $p'_i$ denote the offset-adjusted anchor of query slot $i$, and let $\mathcal{N}_{K_t-1}^{3D}(i)$ denote its $K_t-1$ nearest non-query anchors in Euclidean space. The two support metrics are defined as
\begin{equation}
	D_{\mathrm{geo}}
	=
	\frac{1}{N}
	\sum_{i=1}^{N}
	\sum_{j\in\Omega^{-}(i)}
	\bar{W}_{ij}^{(t)}
	\left\lVert p'_i-p'_j\right\rVert_2
	\label{eq:d_geo}
\end{equation}

\begin{equation}
	S_{\mathrm{geo}}
	=
	\frac{1}{N}	\sum_{i=1}^{N}
	\sum_{j\in\Omega^{-}(i)}
	\bar{W}_{ij}^{(t)}
	\mathbb{I}
	\left[
		j\in\mathcal{N}_{K_t-1}^{3D}(i)
		\right]
	\label{eq:s_geo}
\end{equation}
where $\mathbb{I}[\cdot]$ is the indicator function.

$D_{\mathrm{geo}}$ measures the weighted spatial extent of the non-self support, whereas $S_{\mathrm{geo}}$ measures its weighted overlap with the query anchor's actual 3D nearest-neighbor set. A smaller $D_{\mathrm{geo}}$ and a larger $S_{\mathrm{geo}}$ indicate that OFR assigns more support mass to spatially nearby anchors. Together, the two metrics quantify the geometric compatibility between the local sequence support used by OFR and the underlying point cloud neighborhood.

\begin{table*}[!t]
	\centering
	\caption{Per-category ShapeNetPart results and support analysis. We report IoU$_C$, $D_{\mathrm{geo}}$, and $S_{\mathrm{geo}}$. ``Sinkhorn'' denotes replacing OFR with the Sinkhorn variant.}
	\resizebox{\textwidth}{!}{%
		\begin{tabular}{l|r|ccccccccccccccccccc}
			\toprule
			Method                       & Mean           & \rotatebox{90}{Airpl.} & \rotatebox{90}{Bag} & \rotatebox{90}{Cap} & \rotatebox{90}{Car} & \rotatebox{90}{Chair} & \rotatebox{90}{E-ph.} & \rotatebox{90}{Guitar} & \rotatebox{90}{Knife} & \rotatebox{90}{Lamp} & \rotatebox{90}{Laptop} & \rotatebox{90}{Motor} & \rotatebox{90}{Mug} & \rotatebox{90}{Pistol} & \rotatebox{90}{Rocket} & \rotatebox{90}{Skate} & \rotatebox{90}{Table} \\ \midrule

			\multicolumn{18}{c}{\textit{Category-wise Part IoU$_C$ (\%)}}                                                                                                                                                                                                                                                                                                                                                                             \\ \midrule
			Point-BERT \cite{Point-BERT} & 84.1           & 84.3                   & 84.8                & \textbf{88.0}       & 79.8                & \textbf{91.0}         & \textbf{81.7}         & 91.6                   & 87.9                  & 85.2                 & 95.6                   & 75.6                  & \textbf{94.7}       & 84.3                   & 63.4                   & 76.3                  & 81.5                  \\
			Mamba3D \cite{Mamba3D}       & 84.1           & 84.3                   & 84.0                & 87.9                & 79.8                & 90.9                  & 80.6                  & 91.3                   & 87.2                  & 86.1                 & 95.4                   & 76.1                  & \textbf{94.7}       & 84.8                   & \textbf{64.3}          & 77.6                  & 81.2                  \\
			DM3D                         & \textbf{84.8}  & \textbf{84.9}          & \textbf{85.7}       & \textbf{88.0}       & \textbf{80.6}       & 90.8                  & 81.1                  & \textbf{92.1}          & \textbf{88.7}         & \textbf{86.6}        & \textbf{96.3}          & \textbf{77.9}         & 94.6                & \textbf{85.0}          & 63.6                   & \textbf{78.0}         & \textbf{82.3}         \\
			Sinkhorn \cite{Sinkhorn}     & 84.3           & 84.5                   & 84.8                & 87.5                & 80.2                & 90.5                  & 80.8                  & 91.8                   & 88.2                  & 86.2                 & 95.8                   & 76.6                  & 94.5                & 84.5                   & 63.3                   & 77.6                  & 82.0                  \\ \midrule
			\multicolumn{18}{c}{\textit{Weighted Geometric Distance $D_{\mathrm{geo}}\downarrow$}}                                                                                                                                                                                                                                                                                                                                                    \\ \midrule
			DM3D                         & \textbf{0.216} & 0.187                  & 0.301               & 0.300               & 0.255               & 0.295                 & 0.294                 & \textbf{0.138}         & \textbf{0.124}        & \textbf{0.145}       & \textbf{0.123}         & \textbf{0.194}        & \textbf{0.321}      & \textbf{0.179}         & 0.174                  & 0.157                 & \textbf{0.268}        \\
			Sinkhorn                     & 0.231          & 0.224                  & \textbf{0.300}      & \textbf{0.219}      & \textbf{0.221}      & \textbf{0.269}        & \textbf{0.277}        & 0.225                  & 0.214                 & 0.152                & 0.188                  & 0.229                 & 0.350               & 0.275                  & \textbf{0.135}         & \textbf{0.130}        & 0.294                 \\ \midrule
			\multicolumn{18}{c}{\textit{ Geometric Support Rate $S_{\mathrm{geo}}\uparrow$ (\%)}}                                                                                                                                                                                                                                                                                                                                                     \\ \midrule
			DM3D                         & \textbf{72.5}  & \textbf{76.4}          & \textbf{78.3}       & \textbf{74.2}       & 69.5                & 64.1                  & 70.3                  & \textbf{72.7}          & \textbf{70.3}         & \textbf{71.1}        & \textbf{75.8}          & \textbf{70.3}         & \textbf{75.8}       & 66.4                   & \textbf{77.3}          & \textbf{72.7}         & \textbf{74.2}         \\
			Sinkhorn                     & 71.7           & 74.5                   & 77.2                & 73.0                & \textbf{69.9}       & \textbf{72.4}         & \textbf{72.8}         & 71.0                   & 68.7                  & 69.3                 & 72.7                   & 68.4                  & 74.4                & \textbf{66.5}          & 73.0                   & 71.9                  & 71.7                  \\ \bottomrule
		\end{tabular}%
	}
	\label{tab:ofr_ablation}
\end{table*}

\paragraph{Analysis of OFR feature support}
To examine whether the serialized local window provides geometrically appropriate feature support and how OFR adjusts this support, we visualize representative local windows from the Airplane and Guitar categories in Fig.~\ref{fig:support_weights}.

In panel (a), the yellow path connects the nine tokens contained in the local sequence window. Although these tokens are adjacent in the serialized sequence, several consecutive tokens are separated by relatively long distances in 3D space. As shown by the ground-truth part labels in panel (b), the same window may also span different semantic parts, including the body, wing, and engine of the Airplane, as well as the head and neck of the Guitar. These examples indicate that sequence proximity does not necessarily correspond to spatial proximity or part-level consistency.

Panel (c) visualizes the feature support assigned by OFR. The visual prominence of each highlighted candidate represents its relative support weight. In both examples, OFR assigns larger weights to a small subset of candidates located near the query anchor in 3D, while candidates that are distant from the query or lie across part boundaries receive substantially smaller weights.  OFR therefore changes the feature read at the current sequence slot without moving the token or modifying the base serialization order. Together with the quantitative results of $D_{\mathrm{geo}}$ and $S_{\mathrm{geo}}$ in Table~\ref{tab:ofr_ablation}, the visualization provides an intuitive example of how OFR adjusts local feature support, while the two metrics summarize this behavior over the full dataset.

The per-category results in Table~\ref{tab:ofr_ablation} show that DM3D reaches a mean class IoU of 84.8\%, 0.4\% above the Sinkhorn variant. The largest margins occur for Motorbike (+1.25), Bag (+0.98), and Laptop (+0.51), while the smallest margin is on Mug (+0.07). Relative to Point-BERT and Mamba3D, DM3D improves the category mean by 0.65 and 0.62 points, respectively, and is higher on 12 of 16 categories. This distribution shows that the improvement is not confined to one category.

The support metrics show a similar overall advantage for OFR. DM3D has a lower mean $D_{\mathrm{geo}}$ than the Sinkhorn variant (0.216 vs 0.231), with a reduction in 9 of 16 categories, and a higher mean $S_{\mathrm{geo}}$ (72.5\% vs 71.7\%), with an increase in 12 categories. These results indicate that OFR assigns more weight to candidates that are closer to the current token in 3D.
This is consistent with the goal of improving the features aggregated from a fixed sequence window.
The two metrics are interpreted jointly with the component ablations and qualitative results, thereby providing a more complete assessment of OFR.

\paragraph{Sensitivity to the base serialization}

The choice of base serialization affects the available local context, so we evaluate Hilbert, Z-order, and random orderings. Table \ref{tab:base_serialization} shows that DM3D improves the corresponding base model under every ordering. On ModelNet40 and PB\_T50\_RS, the gains are 1.60 and 2.21 percentage points with Hilbert, 1.87 and 2.20 with Z-order, and 2.29 and 2.32 with random ordering, respectively. DM3D also shows smaller drops when the base order changes from Hilbert to random, declining by 0.60 rather than 1.29 points on ModelNet40 and by 0.80 rather than 0.91 on PB\_T50\_RS. These results indicate that local feature resampling complements the spatial prior supplied by the base traversal without replacing it. Absolute accuracy nevertheless remains dependent on the selected serialization.

\begin{table}[!t]
	\centering
	\footnotesize
	\caption{Impact of base serialization. ``Base'' denotes the corresponding model without the dynamic resampling components.}
	\begin{tabular}{lcccc}
		\toprule
		\multirow{2}{*}{Serialization} & \multicolumn{2}{c}{ModelNet40} & \multicolumn{2}{c}{PB\_T50\_RS}                          \\
		\cmidrule(l){2-3}\cmidrule(l){4-5}
		                               & Base                           & DM3D                            & Base  & DM3D           \\ \midrule
		Hilbert                        & 92.41                          & \textbf{94.01}                  & 88.62 & \textbf{90.83} \\
		Z-order                        & 91.98                          & 93.85                           & 88.25 & 90.45          \\
		Random                         & 91.12                          & 93.41                           & 87.71 & 90.03          \\ \bottomrule
	\end{tabular}
	\label{tab:base_serialization}
\end{table}

\section{Conclusion}
Serialized point cloud models rely on a 1D traversal to process irregular 3D structures, yet neighboring tokens can still draw on features from spatially distant or semantically unrelated regions. In this paper, we presented DM3D, a Dynamic Mamba architecture that retains the base order while adapting the information processed at each sequence slot. Its dynamic path combines spatial and sequence-domain feature resampling with geometry-aware state propagation, paperallowing the model to refine local context and attenuate information carried across large geometric gaps without learning an additional global permutation.

Across classification, few-shot learning, part segmentation, and scene semantic segmentation, DM3D delivers competitive performance on ModelNet40, ScanObjectNN, ShapeNetPart, and S3DIS. It achieves 93.30\% pretrained accuracy on ScanObjectNN PB\_T50\_RS. Its consistent gains over Hilbert, Z-order, and random base serializations show that locally adaptive feature support can complement different traversal priors. Together, DM3D suggests a practical alternative to learning a new order, one that preserves the efficiency and global structure of serialization while allowing local context and state propagation to adapt to point cloud geometry.

\paragraph{Limitations and future work}
In its current form, DM3D is intended to complement rather than replace base serialization. OFR operates within a bounded local window, and its candidate coverage is therefore related to the spatial locality provided by the underlying traversal. CASU likewise uses geometric distance as a lightweight cue for state modulation, and incorporating complementary structural or semantic cues may further broaden its applicability. Future work can explore jointly improving traversal and local adaptation without compromising the efficiency of the current design.


\section*{Acknowledgements}
All authors thank the 512 Lab and 513 Lab of the School of Weapon Science and Technology at Xi'an Technological University.

\section*{CRediT authorship contribution statement}
\textbf{Bin Liu}: Conceptualization, Data curation, Formal analysis, Funding acquisition, Investigation, Methodology, Project administration, Resources, Software, Supervision, Validation, Visualization, Writing – original draft.
\textbf{Chunyang Wang}: Funding acquisition, Supervision.
\textbf{Xuelian Liu}:  Visualization, Validation.
\textbf{Xuemei Li}:  Formal analysis, Writing – original draft
\textbf{Ge Zhang}: Investigation, Supervision.

%
%

\bibliographystyle{elsarticle-num}
\bibliography{main.bib}

\newpage
$ $
\newpage

\appendix
This appendix contains mathematical derivations and supplementary experiments. Section~\ref{Differentiability} establishes the behavior of Offset-Guided Feature Resampling (OFR) for a fixed positive Gaussian scale and then examines its zero- and infinite-scale limits. Section~\ref{More Experimental} reports additional ablations, offset statistics, qualitative examples, and a single-run optimization test that complement the evidence in the main paper.

\section{Analysis of OFR Differentiability}
\label{Differentiability}
In this section, we analyze the Gaussian weights and their derivatives with respect to the offset index values $s_i$ to characterize the behavior of Offset-Guided Feature Resampling (OFR).

OFR restricts the candidate indices for token $i$ to a fixed local index window $\Omega(i)$ centered at its base index $I_i$, with cardinality $K_i = |\Omega(i)| \le K_t$.
The weighting function is defined as:
\begin{equation}\label{weight}
	W_{ij}^{(t)}
	=
	\frac{
		\exp\left(-\frac{(s_i-I_j)^2}{2\sigma_t^2}\right)
	}{
		\sum_{l\in\Omega(i)}
		\exp\left(-\frac{(s_i-I_l)^2}{2\sigma_t^2}\right)
	},
	\quad j\in\Omega(i)
\end{equation}
where $\sigma_t$ is the Gaussian scale parameter in the sequential domain, and $s_i = I_i + \Delta t_i$ denotes the offset index value of the $i$-th token after applying the learned offset $\Delta t_i$.

To analyze gradient propagation, the first-order derivative of $W_{ij}^{(t)}$ with respect to $s_i$ is given by:
\begin{equation}\label{local_derivative}
	\frac{\partial W_{ij}^{(t)}}{\partial {{s}_{i}}}=\frac{W_{ij}^{(t)}}{\sigma _{t}^{2}}\left( {{I}_{j}}-\sum\nolimits_{l\in \Omega (i)}{W_{il}^{(t)}{{I}_{l}}} \right),\quad j\in \Omega (i)
\end{equation}

For $j\notin\Omega(i)$, $W_{ij}^{(t)}=0$ by definition, and thus its derivative with respect to $s_i$ is also zero.
We analyze three cases within the local candidate set $\Omega(i)$: $\sigma_t \to +\infty$, $\sigma_t > 0$, and $\sigma_t \to 0^+$.

\subsection{Case 1: Analysis of $\sigma_t \to +\infty$}
As $\sigma_t \to +\infty$, all exponential terms within the local window $\Omega(i)$ converge to $1$, yielding:
\begin{equation}
	\underset{{{\sigma }_{t}}\to +\infty }{\mathop{\lim }}\,W_{ij}^{(t)}=\frac{1}{{{K}_{i}}} , \quad
	\underset{{{\sigma }_{t}}\to +\infty }{\mathop{\lim }}\,\frac{\partial W_{ij}^{(t)}}{\partial {{s}_{i}}}=0
\end{equation}

In this case, OFR degenerates into local average pooling over the candidate window $\Omega(i)$, and the gradients with respect to $s_i$ vanish.

\subsection{Case 2: Analysis of $\sigma_t>0$}
When $\sigma_t$ is finite and positive, each local weight $W_{ij}^{(t)}$ is infinitely differentiable with respect to $s_i$ for all $j\in\Omega(i)$.
Since $W_{ij}^{(t)}\in(0,1)$ and Eq.~\eqref{local_derivative} is continuous in $s_i$ over the $\Omega(i)$, the derivative is also continuous and remains bounded with respect to $s_i$.

In this case, OFR performs a continuously differentiable local reassignment mapping operation, which supports stable backpropagation through $s_i$.

\subsection{Case 3: Analysis of $\sigma_t \to 0^+$}
According to Eq.~\eqref{weight}, the relative distances between $s_i$ and the candidate indices determine the weights, leading to two scenarios: $s_i$ has a unique nearest candidate, or it is equidistant to two candidates.

Let	${d^2_{\min }}={{\min }_{l\in \Omega (i)}}{{({{s}_{i}}-{{I}_{l}})}^{2}}$ and define the active index set ${{\mathcal{T}}_{i}}=\left\{ j\in \Omega (i)|{{({{s}_{i}}-{{I}_{j}})}^{2}}={d^2_{{{\min }}}} \right\}$ with cardinality $m_i = |\mathcal{T}_i|$. The Gaussian kernel can then be rewritten as:
\begin{equation}
	\begin{aligned}
		\mathcal{W}(s_i-I_j;\sigma_t) & =\exp \left( -\frac{{{({{s}_{i}}-{{I}_{j}})}^{2}}}{2\sigma _{t}^{2}} \right)                                                                                        \\
		                              & =\exp \left( -\frac{{{d}_{\min }}^{2}}{2\sigma _{t}^{2}} \right)\cdot \exp \left( -\frac{{{({{s}_{i}}-{{I}_{j}})}^{2}}-{{d}_{\min }}^{2}}{2\sigma _{t}^{2}} \right)
	\end{aligned}
\end{equation}

For all $j\in {{\mathcal{T}}_{i}}$, the second term equals 1. For $j\notin {{\mathcal{T}}_{i}}$, it decays exponentially. Substituting into the normalized weight yields:
\begin{equation}
	W_{ij}^{(t)}=
	\frac{
	\exp\left(-\frac{(s_i-I_j)^2-d_{\min}^2}{2\sigma_t^2}\right)
	}{
	\sum_{l\in\mathcal T_i}1+
	\sum_{l\notin\mathcal T_i}
	\exp\left(-\frac{(s_i-I_l)^2-d_{\min}^2}{2\sigma_t^2}\right)
	}
\end{equation}

\paragraph{\textbf{Case 3a: $s_i$ has a unique nearest candidate $I_k$ within $\Omega(i)$}}
\label{case3a}
When $s_i$ has a unique nearest local index $I_k$ within $\Omega(i)$, we have $\mathcal{T}_i = \{k\}$ and $m_i = 1$. The non-minimum terms vanish exponentially, yielding:
\begin{equation}
	\underset{{{\sigma }_{t}}\to {{0}^{+}}}{\mathop{\lim }}\,W_{{ij}}^{(t)}=\left\{ \begin{matrix}
		1,j=k    \\
		0,j\ne k \\
	\end{matrix} \right.
\end{equation}

For the derivative, both ${{W}^{(t)}}_{il}$  and its weighted sum decay exponentially, leading to:
\begin{equation}
	\lim_{\sigma_t\to 0^+}\frac{\partial W_{ij}^{(t)}}{\partial s_i}=0, \quad j\in\Omega(i)
\end{equation}

In this scenario, the weight deterministically assigns $s_i$ to its nearest local index $I_k$, and the gradient vanishes, indicating that OFR degenerates into deterministic local reassignment within $\Omega(i)$.

\paragraph{\textbf{Case 3b: $s_i$ is equidistant to two local indices}}

In this scenario, $m_i=2$, and all $j\in \mathcal{T}_i$ have $(s_i-I_j)^2=d^2_{\min}$, while the other indices have larger distances. As ${{\sigma }_{t}}\to {{0}^{+}}$, the minor terms in the denominator vanish, leading to:
\begin{equation}
	\lim_{\sigma_t\to 0^+} W_{ij}^{(t)} =
	\begin{cases}
		\frac{1}{m_i}, & j\in\mathcal{T}_i,    \\
		0,             & j\notin\mathcal{T}_i.
	\end{cases}
\end{equation}

For the derivative, when $j\notin \mathcal{T}_i$,  ${W}^{(t)}_{ij}$ decays exponentially, so the derivative tends to 0. For $j\in \mathcal{T}_i$, the derivative is given by:
\begin{equation}
	\frac{\partial W_{ij}^{(t)}}{\partial {{s}_{i}}}=\frac{1}{{{m}_{i}}\sigma _{t}^{2}}\left( {{I}_{j}}-{{{\bar{I}}}_{{{\mathcal{T}}_{i}}}} \right),\quad {{\bar{I}}_{{{\mathcal{T}}_{i}}}}=\frac{1}{{{m}_{i}}}\sum\limits_{l\in {{\mathcal{T}}_{i}}}{{{I}_{l}}}
\end{equation}
where $\bar I_{\mathcal{T}_i}$ denotes the mean index of $\mathcal{T}_i$. Since $m_i=2$ and the two indices are distinct, $I_j-\bar I_{\mathcal{T}_i}\neq 0$ for each $j\in\mathcal{T}_i$. Therefore:
\begin{equation}
	\underset{{\sigma_t} \to 0^+}{\lim} \frac{\partial W^{(t)}_{ij}}{\partial s_i} =
	\begin{cases}
		0,       & j\notin\mathcal{T}_i,                           \\
		+\infty, & j\in\mathcal{T}_i,\ I_j>\bar I_{\mathcal{T}_i}, \\
		-\infty, & j\in\mathcal{T}_i,\ I_j<\bar I_{\mathcal{T}_i}
	\end{cases}
\end{equation}

In this scenario, the weights are evenly distributed across the equidistant local candidates, while the derivatives with respect to $s_i$ diverge. However, even a slight perturbation of $s_i$ breaks the symmetry and restores finite gradients, as in Case 3a. Consequently, the model is unlikely to remain in such equidistant states during training.

Overall, OFR maps each offset index value  $s_i$ to a probabilistic assignment over the local candidate window $\Omega(i)$, which provides a continuous and differentiable relaxation of local discrete reassignment in the index space.
For any finite $\sigma_t>0$, the weighting function remains smooth and differentiable with respect to $s_i$, enabling stable gradient propagation.
As $\sigma_t\to 0^+$, the mapping converges to nearest-candidate selection within $\Omega(i)$, i.e., deterministic local reassignment.
Conversely, as $\sigma_t \to \infty$, OFR degenerates into local average pooling.
Hence, ${{\sigma }_{t}}$  is typically chosen to be relatively small in practice.
Section~\ref{More Experimental} reports the observed behavior of this parameter in training.

\section{More Experimental Results}
\label{More Experimental}

\subsection{Additional Ablation Studies}
\paragraph{\textbf{Candidate reuse and resampling neighborhood size $K_r$}}
The spatial resampling step is designed to refine feature support within the local region identified by the initial ball query rather than rebuild the neighborhood graph after every predicted offset. Table~\ref{ks} first compares these two choices at $K_r=4$. Re-querying after applying $\Delta p$ obtains 90.74\%/89.05\% on OBJ\_ONLY/PB\_T50\_RS, whereas reusing the original candidates reaches 91.74\%/90.83\%. The 1.00/1.78-point difference favors keeping a stable local candidate set while adapting its feature weights. This comparison does not isolate whether the loss comes from changed neighbors or from the additional query-and-gather operation, but it provides no accuracy benefit for rebuilding the neighborhood in the evaluated setting. We therefore reuse the initial candidates in the reported model.

Among the reused-neighborhood settings, increasing $K_r$ from 2 to 4 improves accuracy from 90.96\%/89.46\% to 91.74\%/90.83\%. Increasing it further to 5 lowers the scores to 91.22\%/90.48\%. This pattern is consistent with a trade-off between candidate coverage and overly broad aggregation. We therefore use $K_r=4$ for both the offset cue and spatial resampling.

\begin{table}[!t]
	\centering
	\footnotesize
	\caption{Ablation on candidate reuse and neighborhood size $K_r$ in LCFA and DSR.}
	\begin{tabular}{cccc}
		\toprule
		Reuse neighbors            & $K_r$ value & OBJ\_ONLY      & PB\_T50\_RS    \\ \midrule
		\ding{55}                  & 4           & 90.74          & 89.05          \\ \midrule
		\multirow{4}{*}{\ding{51}} & 2           & 90.96          & 89.46          \\
		                           & 3           & 91.18          & 89.5           \\
		                           & 4           & \textbf{91.74} & \textbf{90.83} \\
		                           & 5           & 91.22          & 90.48          \\ \bottomrule
	\end{tabular}
	\label{ks}
\end{table}

\paragraph{\textbf{Reliability of neighborhood reuse}}
The comparison above evaluates the accuracy effect of neighborhood reuse. To separately examine whether the learned deformation remains local, we report the spatial offset magnitudes $\|\Delta p\|_2$ relative to the ball-query radius $r$ in Table ~\ref{tab:delta_p_stats}. The statistics use 256 samples per dataset and 128 points per sample, yielding 32,768 offsets for each dataset.

The 95th percentile of $\|\Delta p\|_2$ is below $r$ on all three datasets, and fewer than 1\% of offsets have magnitude greater than $r$. These measurements show that the learned anchor displacement is usually smaller than the query radius, which supports the intended locality of the deformation. Offset magnitude alone does not prove that a deformed anchor has exactly the same nearest neighbors; the re-querying ablation in Table ~\ref{ks} provides the corresponding task-level check.

\begin{table}[!t]
	\centering
	\footnotesize
	\caption{Statistics of the spatial offset magnitudes $\|\Delta p\|_2$ with respect to the ball query radius $r$.}
	\begin{tabular}{cccc}
		\toprule
		Dataset             & ModelNet40 & OBJ\_ONLY & PB\_T50\_RS \\ \toprule
		Query radius ($r$)  & 0.2        & 0.25      & 0.25        \\
		Mean value          & 0.118      & 0.130     & 0.121       \\
		${95}$th percentile & 0.162      & 0.200     & 0.179       \\
		Violation count     & 180        & 124       & 26          \\
		Violation rate      & 0.55\%     & 0.38\%    & 0.08\%      \\ \bottomrule
	\end{tabular}
	\label{tab:delta_p_stats}
\end{table}

\paragraph{Local-window size}
The candidate-window size $K_t$ controls the number of local source slots available to OFR. In Table ~\ref{tab:kt}, accuracy increases from $K_t=5$ to $K_t=9$ on both splits; $K_t=11$ changes PB\_T50\_RS only slightly but lowers OBJ\_ONLY. We therefore use $K_t=9$.

\begin{table}[!t]
	\centering
	\footnotesize
	\caption{Sensitivity to the OFR local-window size $K_t$.}
	\begin{tabular}{ccc}
		\toprule
		$K_t$ & OBJ\_ONLY (\%) & PB\_T50\_RS (\%) \\ \midrule
		5     & 91.49          & 90.31            \\
		7     & 91.54          & 90.77            \\
		9     & \textbf{91.74} & \textbf{90.83}   \\
		11    & 91.37          & 90.80            \\ \bottomrule
	\end{tabular}
	\label{tab:kt}
\end{table}

\begin{table}[!t]
	\centering
	\footnotesize
	\caption{Comparison of path-integration operators.}
	\begin{tabular}{clccc}
		\toprule
		 & Variants                & OBJ\_ONLY      & PB\_T50\_RS    & \\ \midrule
		 & Element-wise mean       & 90.88          & 89.02          & \\
		 & Convolutional fusion    & 90.01          & 87.93          & \\
		 & Path integration (Ours) & \textbf{91.74} & \textbf{90.83} & \\ \bottomrule
	\end{tabular}
	\label{fusion}
\end{table}

\paragraph{Choice of path-integration operator}
Table~\ref{fusion} compares the operator used to combine the dynamic and standard paths. Ours exceeds element-wise averaging by 0.86/1.81 percentage points and convolutional fusion by 1.73/2.90 points on OBJ\_ONLY/PB\_T50\_RS. These single-run comparisons support the learned cross-path modulation used in the complete architecture, but do not by themselves establish why it performs better or whether it is optimal among all fusion designs.

\begin{figure}[!t]
	\centering
	\includegraphics[width=\linewidth]{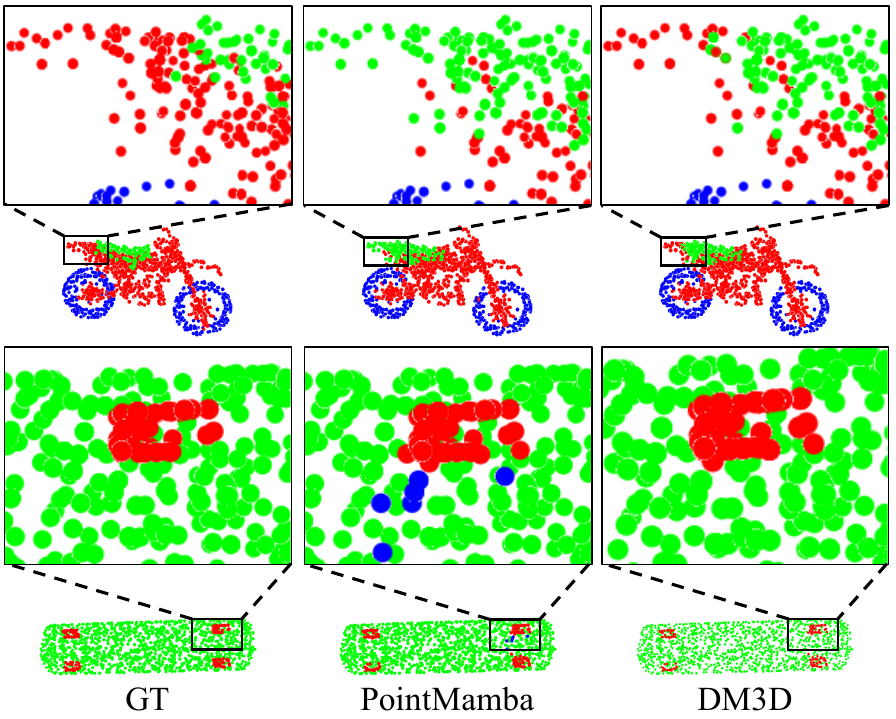}
	\caption{Detailed visualizations of part segmentation results.}
	\label{fig_partseg2}
\end{figure}

\subsection{Part Segmentation Visualization}
Fig. ~\ref{fig_partseg2} presents qualitative comparisons on the ShapeNetPart \cite{ShapeNet} test set, including Ground Truth (GT), PointMamba \cite{PointMamba} predictions, and our DM3D predictions.
In the displayed examples, DM3D produces cleaner part separation and fewer visibly misclassified points than PointMamba. The enlarged motorbike view shows less confusion between the seat and adjacent body regions, while the skateboard example shows more compact wheel predictions with fewer isolated points.

\subsection{Optimization Behavior of OFR}
\begin{figure}[t]
	\centering
	\includegraphics[width=2.2in]{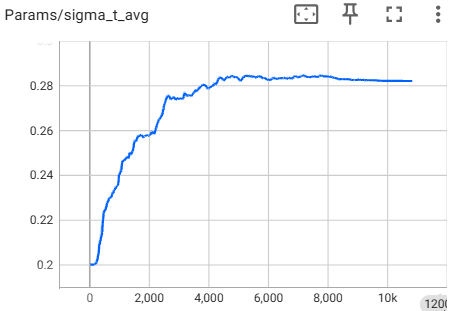}
	\hfil
	\includegraphics[width=2.2in]{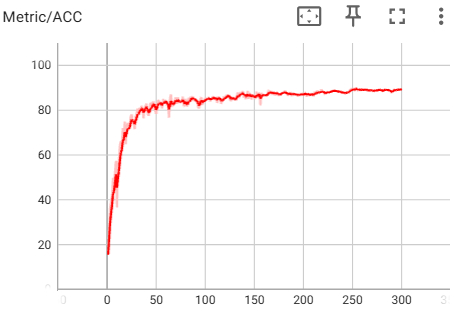}
	\caption{Optimization traces for the D-SSM-only configuration by TensorBoard. Top: learned ${\sigma}_{t}$ over training steps. Bottom: overall accuracy over 300 epochs. Both panels report the same single run.}
	\label{D_SSM_acc}
\end{figure}

Section~\ref{Differentiability} shows that the local Gaussian weights are differentiable with respect to the sampling location $s_i$ when $\sigma_t$ is finite and positive. To complement this analysis, we examine the empirical evolution of the learnable scale parameter $\sigma_t$ in the reported D-SSM-only run. As shown in Fig.~\ref{D_SSM_acc}, $\sigma_t$ is initialized to 0.2 and gradually converges to approximately 0.28 over 300 epochs while remaining positive throughout training. In the same run, the accuracy curve remains stable, and no NaNs or optimization collapse are observed.

\end{document}